\documentclass[10pt,twocolumn,letterpaper]{article}

\usepackage{3dv}
\usepackage{times}
\usepackage{epsfig}
\usepackage{graphicx}
\usepackage{amsmath}
\usepackage{amssymb}
\usepackage[pagebackref=true,breaklinks=true,colorlinks,bookmarks=false]{hyperref}

\usepackage{multirow}
\usepackage{float}
\usepackage{array}

\usepackage{pdfpages}
\usepackage{multido}
\usepackage{wrapfig}
\usepackage[table]{xcolor}
\definecolor{LightCyan}{rgb}{0.88,1,1}
\definecolor{LightYellow}{rgb}{1,1,0.7}
\usepackage{verbatim}
\usepackage[percent]{overpic}
\usepackage{bbding}
\usepackage{pifont}
\usepackage{wasysym}
\usepackage{pifont}
\usepackage{xcolor}
\usepackage{bm}

\newcommand{\gt}{ground-truth}
\definecolor{ours}{rgb}{1,1,0.7}
\definecolor{lower}{RGB}{232, 161, 148}

\usepackage{mathtools}

\definecolor{lower}{RGB}{232, 161, 148}

\newcommand\blfootnote[1]{%
  \begingroup
  \renewcommand\thefootnote{}\footnote{#1}%
  \addtocounter{footnote}{-1}%
  \endgroup
}

\threedvfinalcopy 

\ifthreedvfinal\fi
\begin{document}

\title{Neural Disparity Refinement for Arbitrary Resolution Stereo}

\author{Filippo Aleotti$^*$ \hspace*{1cm} Fabio Tosi$^*$ \hspace*{1cm} Pierluigi Zama Ramirez$^*$ \\
Matteo Poggi \hspace*{1cm} Samuele Salti \hspace*{1cm} Stefano Mattoccia  \hspace*{1cm} Luigi Di Stefano \\
CVLAB, Department of Computer Science and Engineering (DISI)\\
University of Bologna, Italy\\
{\tt\small \{filippo.aleotti2, fabio.tosi5, pierluigi.zama\}@unibo.it}
}

\twocolumn[{
\renewcommand\twocolumn[1][]{#1}
\maketitle
\begin{center}
\vspace{-0.5cm}
\includegraphics[width=0.7\linewidth]{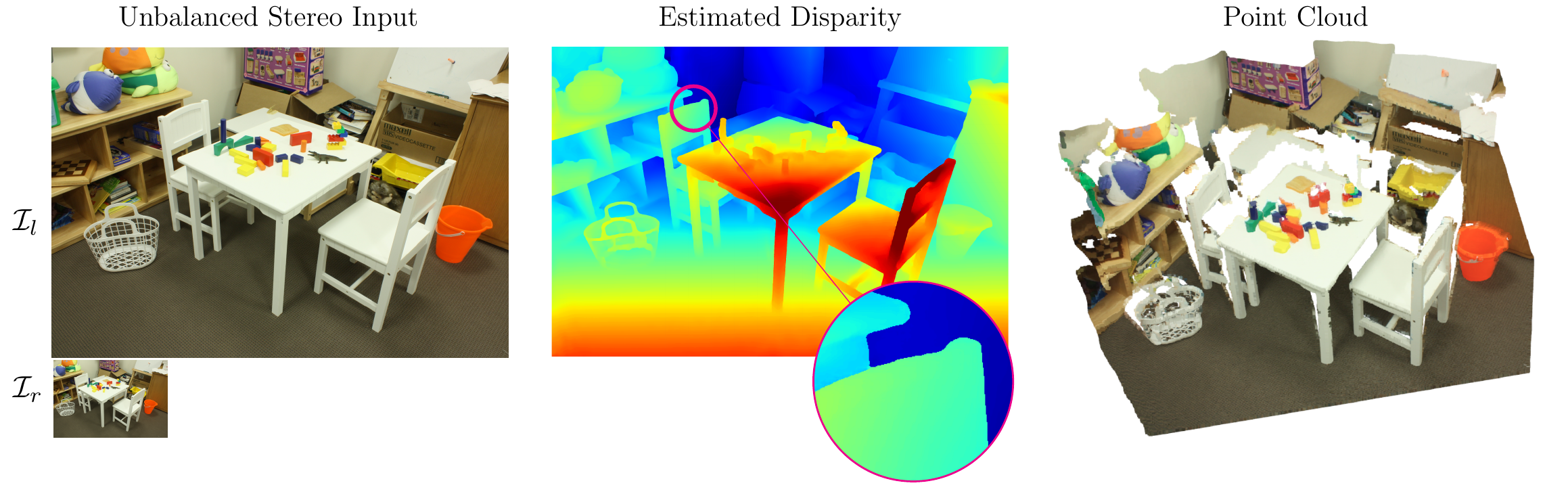}
\label{fig:teaser}
\end{center}
\vspace{-0.5cm}
\small \hypertarget{fig:teaser}{Figure 1.} \textbf{Example of Arbitrary Resolution Stereo.} Given an \textbf{unbalanced} stereo pair made of a high-res image, $\mathcal{I}_l$, with shape $2724\times1848$ and a low-res image, $\mathcal{I}_r$, at $691\times462$,  both unseen and real, our method can estimate a high-res disparity map at $2742\times1848$ with sharp edges (leading to a clean point cloud near depth discontinuities) based on a single  \textbf{training on synthetic data only}.
\vspace{0.4cm}
}]

\maketitle

\begin{abstract}

We introduce a novel architecture for neural disparity refinement aimed at facilitating deployment of 3D computer vision  on cheap and widespread consumer devices, such as mobile phones. Our approach relies on a continuous formulation that enables to estimate a refined disparity map at any arbitrary output resolution. Thereby, it  can handle effectively the
unbalanced camera setup typical of nowadays mobile phones, which feature both  high and low resolution RGB sensors within the same device. Moreover, our neural network can  process seamlessly the output of a variety of stereo methods and, by refining the disparity maps computed by a traditional matching algorithm like SGM, it can  achieve unpaired zero-shot generalization performance compared to state-of-the-art end-to-end stereo models. 

\end{abstract}

\section{Introduction}
\blfootnote{$^*$ Joint first authorship.}Depth perception from images can be effectively deployed by mobile agents, such as vehicles and robots, to navigate and interact with our 3D world more and more autonomously.  
Stereo vision \cite{scharstein2002taxonomy,poggi2021synergies} is among the prominent methodologies to infer depth from images due to easiness of deployment and good accuracy.  Leveraging on large collections of image pairs annotated with ground-truth disparity labels \cite{Geiger2012CVPR,Menze2015CVPR,scharstein2014high,schoeps2017cvpr},  the latest advances in stereo  mostly rely on deep learning. In particular, end-to-end models \cite{zhang2019ga,cheng2020hierarchical} can reach unpaired accuracy if evaluated in the same domain as that on which they are trained.   
Nowadays, the ever increasing availability of multiple cameras on consumer devices  -- \eg mobile phones -- paves the way to leverage on stereo for 3D applications on a variety of relatively cheap and massively widespread platforms. Yet, we argue that, on the road toward unconstrained deployment on consumer devices, state-of-the-art stereo solutions may need to face two main challenges concerning, on one hand, the peculiar camera configuration, on the other the ability to generalize to unseen domains. 
As for the former, most modern mobile devices implement an \textit{unbalanced} camera setup, often consisting of a high-resolution sensor -- up to dozens of Megapixels (Mpx) -- coupled with cheaper, lower-resolution cameras of a few Mpx. This kind of configuration, which differs from the classical \textit{balanced} stereo setup, has been studied in literature only recently and in simplified settings  \cite{Liu_2020_CVPR},  \ie  with the resolution of the larger image being way lower than 1Mpx. 
The latter, indeed a challenge for any sort of learning machinery, has started to be investigated more thoroughly in the stereo literature in the past couple of years \cite{zhang2019domaininvariant,cai2020matchingspace,Tankovich_2021_CVPR}. 

In this paper, we introduce a novel framework which can tackle both the challenges highlighted above. Inspired by traditional,  pre-deep learning stereo pipelines \cite{scharstein2002taxonomy}, we focus on the last step and design a refinement module guided by the reference RGB image of the input stereo pair which is capable of enhancing both the quality and  resolution of an input disparity map. Our module, implemented as a neural network, is general and, after being trained once and solely on synthetic data, can be used to refine a disparity map produced by any blackbox stereo method, either a traditional \cite{hirschmuller2007stereo} or learned \cite{luo2016efficient}  matcher  as well as an end-to-end  network \cite{chang2018psmnet,yang2019hierarchical}. Thanks to a continuous formulation, our neural refinement module can predict disparity at any arbitrary location of the image space. This is conducive to estimate the output disparity map at any  desired resolution and  realize  an effective and elegant approach to process unbalanced stereo images.
To address the generalization issue of learning-based stereo, we propose to deploy a traditional, domain-agnostic stereo algorithm in order to provide the input disparity map to our neural refinement module. Thereby, we realize an hybrid handcrafted-learned solution that neatly outperforms state-of-the-art deep networks when processing previously unseen image content, a setting referred to in literature as zero-shot generalization. 
Moreover, the \emph{refined} disparity maps predicted by our framework are sharp at depth discontinuities thanks to a novel loss function that allows for expressing the output disparities as a combination of a categorical value and a continuous offset. Exhaustive experimental results on a large variety of datasets support the following main claims.

\begin{itemize}

    \item Our neural module significantly outperforms existing deep refinement approaches \cite{gidaris2017detect,batsos2018recresnet,ferrera2019fast} when processing raw disparity maps from off-the-shelf stereo matchers. Moreover, unlike other proposals, we demonstrate the ability to improve disparity maps computed by end-to-end  stereo networks. 
    
     \item The versatility of our architecture, which can handle any arbitrary output resolution,  allows for dealing effectively with unbalanced stereo images, outperforming the accuracy of end-to-end models when deployed for this task.
    
    \item When combined with traditional stereo algorithms, our disparity refinement approach achieves superior accuracy in zero-shot generalization to unseen domains compared to state-of-the-art stereo networks  \cite{cai2020matchingspace,zhang2019domaininvariant} without penalizing in-domain performance.
    
    \item Our novel formulation concerning the computation of  output disparities yields sharp  maps at depth discontinuities, which  results in more accurate estimations compared to other existing output representations \cite{Tosi2021CVPR} and clean 3D reconstructions. 
    
    \end{itemize}

The main contributions provided by our proposal are summarized visually in Fig. \hyperref[fig:teaser]{1}.  

\section{Related Works}
We briefly review the literature relevant to our work.

\textbf{Traditional Stereo.} Stereo matching has a longstanding history in computer vision. Most hand-made algorithms perform a subset of the steps outlined in \cite{scharstein2002taxonomy} and are classified, accordingly, into local or global strategies. The former approaches are usually fast and na\"ively look for matches across local patches according to a robust function \cite{Secaucus_1994_ECCV}, while the latter involve complex optimization schemes at the cost of much higher runtime \cite{yamaguchi2012continuous}.
A trade-off between accuracy and speed is obtained by the Semi-Global matching (SGM) algorithm \cite{hirschmuller2007stereo}. The very first attempts to improve stereo with deep learning focused on replacing the matching cost functions with neural networks \cite{zbontar2016stereo,Chen_2015_ICCV,luo2016efficient} within the SGM pipeline.

\textbf{Disparity Refinement.} The last step of traditional stereo pipelines \cite{scharstein2002taxonomy} usually involves refining the estimated disparities, which are affected by errors especially near occlusions, thin structures or reflective surfaces. These artifacts can be not only localized \cite{poggi2021confidence}, but also corrected \cite{tosi2017learning,kim2017deep}. In the past, the refinement task has been faced using non local-means filtering \cite{favaro2010recovering}, dictionary-based strategies \cite{kwon2015data} or filter forests \cite{fanello2014forestfilter}. More recent refinement approaches rely on deep networks, for instance to sequentially detect, replace and refine noisy pixels \cite{gidaris2017detect}. Batsos and Mordohai \cite{batsos2018recresnet} cast refinement as a recurrent process performed by a neural network over the disparity map, while Jie \etal \cite{Jie_2018_CVPR} recurrently perform a left-right consistency check. Finally, Ferrera \etal \cite{ferrera2019fast} fuse disparity maps from several stereo algorithms with the one estimated by a neural network.

\begin{figure*}[!ht]
\centering
\includegraphics[trim=0cm 3.8cm 0cm 5.7cm,clip,width=0.85\textwidth]{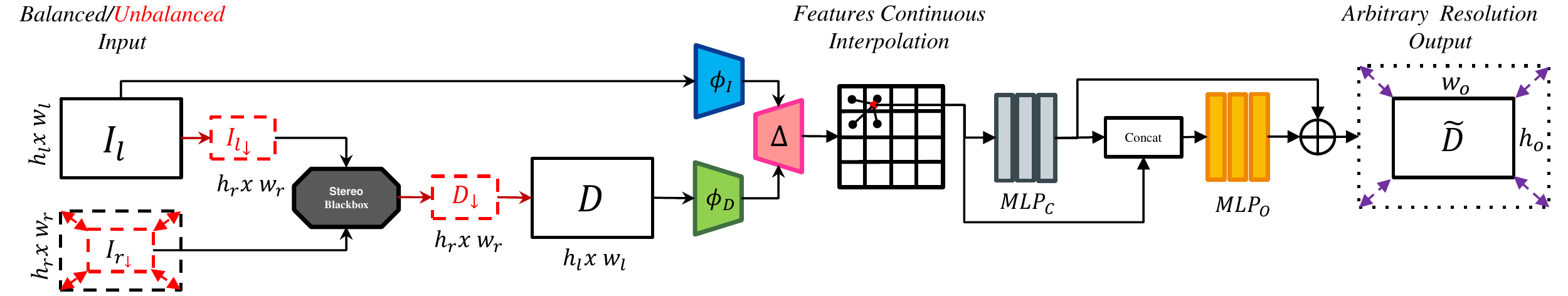}
\caption{\textbf{Neural Disparity Refinement, architecture overview.} Given a rectified stereo pair captured using either a balanced or unbalanced (red dotted lines) stereo setting, our goal is to estimate a refined disparity map $\widetilde{\mathcal{D}}$ at any arbitrary spatial resolution starting from noisy disparities $\mathcal{D}$ pre-computed by any existing stereo blackbox. We first extract deep high-dimensional features from $\mathcal{I}_l$ and $\mathcal{D}$  using two separate convolutional branches, $\phi_\mathcal{I}$ and $\phi_\mathcal{D}$, that are combined together by a  decoder, $\Delta$. Then, at each continuous 2D location in the $\mathcal{I}_l$ image domain, we interpolate features across the levels of  $\Delta$ in order to feed them into a disparity estimation module realized through two MLPs, namely $\textrm{MLP}_C$ and $\textrm{MLP}_O$, which predict an integer disparity value and a sub-pixel offset,  respectively.}
\label{fig:network}
\end{figure*}

\textbf{Deep Stereo.} Nowadays the most popular trend consists, by far, in training end-to-end deep neural networks \cite{poggi2021synergies}. The seminal work by Mayer \etal \cite{mayer2016large} represented a turning point in the stereo literature, proposing the first deep architecture (DispNet) alongside a large synthetic dataset, Freiburg SceneFlow, to pre-train it. Follow-up works improved over DispNet, for instance by explicitly building a feature-based cost volume \cite{Kendall_2017_ICCV} optimized by means of 3D convolutions and leading to two main families of deep stereo networks based on either  2D   \cite{Pang_2017_ICCV_Workshops,Liang_2018_CVPR,Ilg_2018_ECCV} or 3D \cite{chang2018psmnet,zhang2019ga,yang2019hierarchical,cheng2020hierarchical} convolutions. In both cases, some works deployed multi-task frameworks combining stereo with semantic segmentation \cite{yang2018segstereo,dovesi2020real}, edge detection \cite{song2018edgestereo} and optical flow \cite{Jiang_2019_ICCV,aleotti2020learning,lai19cvpr,wang2019unos} estimation. 
Usually, the training procedure shared by most stereo networks consists in a pre-training phase on synthetic data (\eg SceneFlow \cite{mayer2016large}) to initialize the model followed by fine-tuning on a real dataset, such as KITTI \cite{Geiger2012CVPR}. The latter step is usually necessary to overcome the domain shift occurring between computer-generated imagery (CGI) and real images.
To tackle this problem, two orthogonal families of approaches recently arose in literature. The first relies on self-supervision \cite{Tonioni_2017_ICCV,tonioni2020unsupervised,Zhou_2017_ICCV,wang2019unos,lai19cvpr,aleotti2020reversing} to either directly train on real images without requiring ground-truth disparities or even keep  adapting the model throughout its actual deployment  \cite{Tonioni_2019_CVPR,Tonioni_2019_learn2adapt,Poggi2021continual}.
The second family aims at training neural networks that are robust to domain shifts even if trained on synthetic images only. This trend has been recently explored adopting domain invariant batch normalization \cite{zhang2019domaininvariant} or replacing RGB features encoders with classical matching functions invariant to specific images transformations \cite{cai2020matchingspace}.
Finally, another challenging and rarely explored topic in deep stereo concerns high resolution images. In fact, nowadays,  HSMNet \cite{yang2019hierarchical} is the sole architecture that can process full resolution Middlebury images with good accuracy, while SMD-Nets \cite{Tosi2021CVPR} allows to infer high-resolution disparity maps starting from lower resolution stereo images. The unbalanced setting discussed in the introduction has been explored only by Liu \etal \cite{liu2020visually}  by conducting experiments on the KITTI dataset \cite{Menze2015CVPR} with images having an average resolution lower than 0.5 Mpx.

\textbf{Continuous Output Representation}. Nowadays, implicit neural representations exhibit undisputed results in many computer vision tasks such as image synthesis \cite{Mildenhall2020ECCV,Schwarz2020NIPS}, 3D reconstruction \cite{Mescheder2019CVPR,Park2019CVPR,Chen2019CVPR,Saito2019ICCV,Sitzmann2020NIPS,Niemeyer2020CVPR,Peng2020ECCV} and semantic segmentation \cite{Kirillov2020CVPR}. Very recently, continuous functions have been also adopted in the stereo matching field \cite{Tosi2021CVPR} allowing to predict disparity at any continuous pixel location. Differently from \cite{Tosi2021CVPR}, we adopt a different output representation to estimate the final disparity map.

\section{Proposed Architecture}
In this section, we introduce our Neural Disparity Refinement architecture that, given an input disparity map computed by any black-box stereo method, allows for refining it at any desired output resolution, \eg \emph{higher} than the input one. 
To this aim, we propose a simple yet effective neural architecture that accepts as inputs  the reference RGB image of a stereo pair alongside a  corresponding noisy disparity map, the latter possibly even at a lower resolution than the former. A standard convolutional neural network extracts and combines deep features computed from both inputs. Then, the final features are feed into two Multi-Layer Perceptrons (MLPs) that, thanks to a continuous formulation, allows for estimating a refined disparity map at any chosen resolution. The whole network is trained end-to-end based on  a novel loss function that enables to predict sharp and precise disparities at object boundaries. In the remainder of this section we explain in detail the key components of our proposed architecture.

\subsection{Continuous Disparity Refinement Network}

Our network implements three different steps: i) feature extraction, ii) feature interpolation and iii) disparity prediction with subpixel refinement, as illustrated in Figure \ref{fig:network}.

\textbf{Feature Extraction.} 
Given two rectified stereo images, $\mathcal{I}_l$ and $\mathcal{I}_r$, with shapes $w_l \times h_l$ and $w_r \times h_r$ and the same aspect ratio ($\frac{w_l}{h_l}=\frac{w_r}{h_r}$), we aim at obtaining a refined disparity map, $\widetilde{\mathcal{D}}$, at any arbitrary spatial resolution $w_o \times h_o$. Depending on factor 
$\kappa=\frac{w_l}{w_r}$, 
we may have a \textit{balanced} ($\kappa=1$) or \textit{unbalanced} ($\kappa \neq 1$) stereo setup. In the latter one, the rectification constraint shall be understood to hold up to a scale factor, \ie $\mathcal{I}_l$ and $\mathcal{I}_r$ turn out to be rectified whenever resized to the same -arbitrary- shape.  Hereinafter, we will also refer to 
$\kappa$  as to \emph{unbalance} factor.

The disparity map to be refined,  $\mathcal{D}$, may come from any stereo approach, either traditional or learned. In both the balanced and unbalanced setups we assume $\mathcal{D}$ to have the same resolution as the reference stereo image $\mathcal{I}_l$. In particular, as depicted in Figure \ref{fig:network}, in case of  unbalanced setup,  we assume that a low-resolution disparity map $\mathcal{D}_{\downarrow}$ is computed by the stereo blackbox by downsampling $\mathcal{I}_l$ to the same resolution as $\mathcal{I}_r$  and later upsampled to match the shape of $\mathcal{I}_l$. Then, two separate convolutional encoders, $\phi_{\mathcal{I}}$ and $\phi_{\mathcal{D}}$, extract features at different resolutions, collectively denoted here as  $\boldsymbol{\mathcal{F}}_{\mathcal{I}}$ and $\boldsymbol{\mathcal{F}}_{\mathcal{D}}$,  from $\mathcal{I}_l$ and $\mathcal{D}$, respectively. A decoding stage, referred to as $\Delta$ in Figure \ref{fig:network},  is in charge of merging the features from the two encoders while restoring the original $\mathcal{I}_l$ resolution. At each level $l$ of the decoder $\Delta$, the features from the corresponding encoder levels, $\boldsymbol{\mathcal{F}}^l_{\mathcal{I}}$ and $\boldsymbol{\mathcal{F}}^l_{\mathcal{D}}$, are aggregated (by mean of channel-wise sum) and used as skip-connections for the upsampled features from the previous decoder level, denoted here as $\boldsymbol{\mathcal{F}}^{l-1}_{\Delta}$.

\textbf{Feature Interpolation.}
In order to infer the refined disparity map $\mathcal{\widetilde{D}}$ at any arbitrary resolution, similarly to \cite{Tosi2021CVPR}, we formulate the disparity prediction problem as the estimation of a function defined on a continuous 2D domain. In particular, rather than directly predicting a disparity map from $\Delta$, first we  compute features across  the decoder levels, $\boldsymbol{\mathcal{F}}^{l}_{\Delta}$, at any arbitrary continuous location of the 2D image domain $\mathcal{I}_l$ by bilinearly interpolating between the four nearest discrete locations.  
Then, the interpolated features computed at each level are concatenated and forwarded to a  Multi-Layer Perceptron (MLP) that provides the disparity prediction, $\widetilde{\mathcal{D}}$, at the considered continuous 2D location. As the MLP predicts a disparity value based on the corresponding point-wise features, the proposed formulation allows for choosing any desired output resolution by simply sampling features at continuous spatial locations of the 2D domain.

\textbf{Disparity Prediction with Subpixel Precision.} Similar to standard regression tasks, existing disparity refinement approaches  adopt a  $\mathcal{L}_1$ loss \cite{gidaris2017detect, batsos2018recresnet, ferrera2019fast}. However, this choice can cause severe over-smoothing effects  at depth discontinuities \cite{chen2019over,Tosi2021CVPR}, which result in bleeding artifacts when pixels are converted into a 3D point cloud. This problem, which may impact quite negatively on the downstream 3D application, is typically caused by the multi-modal disparity distributions occurring at object boundaries and the smooth function approximation yielded by standard neural network. Very recently, the over-smoothing effect has been tackled in the stereo literature by forcing the multi-modal distribution to be uni-modal \cite{chen2019over} or by predicting  bimodal mixture densities \cite{Tosi2021CVPR} aimed at modelling both the foreground and background disparities near edges. As an alternative, we leverage a simple yet effective strategy that allows us to alleviate the over-smoothing effect as well as to achieve accurate  disparity estimations. In particular, as shown in Figure \ref{fig:network},  for the given continuous location in the 2D domain, we deploy  i) a first MLP, denoted as $\textrm{MLP}_C$, to predict a categorical disparity distribution by casting disparity estimation as a classification task and ii) a second MLP, namely $\textrm{MLP}_O$, to regress a    sub-pixel offset which is added to the most likely integer disparity  predicted by $\textrm{MLP}_C$. As depicted in Figure \ref{fig:network},   both the MLPs process the features computed at the given spatial location, referred to in the following as $\boldsymbol{\mathcal{F}}_\Delta$, these being further concatenated with the predicted integer disparity in order to provide the whole input to the second one. Thus, Equation \ref{eq:disparity} describes how the disparity $\widetilde{\mathcal{D}}$ is predicted by our model at any given 2D location in the image:

\begin{equation}\label{eq:disparity}
  \begin{split}
    \widetilde{\mathcal{D}} =  \textrm{argmax}(\textrm{MLP}_C(\boldsymbol{\mathcal{F}}_\Delta)) + \textrm{MLP}_O(\boldsymbol{\mathcal{F}}_\Delta) \\
    \end{split}
\end{equation}

 where $\textrm{MLP}_C$ is trained to predict a discrete disparity by a cross-entropy loss, while $\textrm{MLP}_O$ predicts an offset in the range $[-1,1]$ and is trained using an $\mathcal{L}_1$ loss.
Accordingly, the final activations of $\textrm{MLP}_C$ and $\textrm{MLP}_O$ are  the $\textrm{Softmax}$ and $\textrm{Tanh}$ functions, respectively.
Denoting $\mathcal{D}^*$ as the ground-truth disparity, the final loss function, $\mathcal{L}$, is computed as the sum of two terms:

\begin{equation} \label{eq:loss}
    \begin{split}
        \mathcal{L} = &-\mathcal{N}(\mathcal{D}^*,\sigma) * \textrm{log}\Big( \textrm{MLP}_C(\boldsymbol{\mathcal{F}}_\Delta) \Big) \\
        &+ \Big| \textrm{MLP}_O(\boldsymbol{\mathcal{F}}_\Delta)-\mathcal{D}^*_s \Big| \\
    \end{split}
\end{equation}
where $\mathcal{N}(\mathcal{D}^*,\sigma)$ is a Gaussian distribution centered at $\mathcal{D}^*$ ($\sigma=\sqrt{2}$ in our experiments), while $\mathcal{D}^*_s$ denotes the difference between $\mathcal{D}^*$ and the integer disparity predicted by $\textrm{MLP}_C$, respectively:
  
\begin{align}
    \mathcal{D}^*_s &= \mathcal{D}^* - \textrm{argmax}\Big(\textrm{MLP}_C(\boldsymbol{\mathcal{F}}_\Delta)\Big)
\end{align}
The latter term in Equation \ref{eq:loss} is minimized only if $\mathcal{D}^*_s$ results in $[-1,1]$.

\section{Experiments}

In this section, we first describe the datasets adopted to validate our proposal. Then, we carry out extensive experiments to demonstrate the benefits brought in by the proposed neural disparity refinement approach when addressing  both balanced as well as unbalanced stereo settings.

\subsection{Datasets}
\textbf{SceneFlow.} The SceneFlow dataset \cite{MIFDB16} is a widely adopted synthetic dataset containing around $35k$ low-resolution ($960\times540$) stereo pairs with dense \gt{} disparity maps. In our experiments, we use 22340 stereo pairs for training, 50 for validation and 387 for testing. 

\textbf{KITTI.} The KITTI dataset \cite{Menze2015CVPR} is a low-res ($\sim$ 0.4 Mpx) real-world stereo dataset  depicting driving scenarios. We rely on KITTI 2012 \cite{Geiger2012CVPR} (194  training  and  195 testing stereo pairs) and KITTI 2015 \cite{Menze2015CVPR} (200  training  and  200 testing pairs) versions, providing sparse  ground-truth  disparities.

\textbf{Middlebury v3.} The Middlebury v3 \cite{scharstein2014high} is a popular high-resolution ($\sim$ 5 Mpx) real-world stereo dataset depicting indoor scenarios and featuring  accurate \gt{} disparity maps, provided at full-res (F) or downsampled to half (H) and quarter resolution (Q).

\textbf{ETH3D.} The ETH3D dataset \cite{schops2017multi} includes a mix of indoor and outdoor scenes,  counting 27 grayscale low-res ($\sim$ 0.4 Mpx) stereo pairs with \gt{} disparity maps.

\textbf{UnrealStereo4K.} The UnrealStereo4K is a realistic synthetic high-resolution (3840$\times$2160) stereo dataset with available \gt{} disparities. Following \cite{Tosi2021CVPR}, we use 7720 images for training, 80 for validation and 200 for testing. Given its high resolution, we employ this dataset in  the unbalanced setup as it allows to simulate different $\kappa$ factors.

\subsection{Implementation Details}
\label{sec:implementation_details}

Our framework is implemented in PyTorch \cite{paszke2019pytorch} and it  is trained using a single NVIDIA 3090 GPU.  All modules are initialized from scratch. We use Adam \cite{kingma2015adam} with $\beta_1= 0.9$  and $\beta_2= 0.999$.  We adopted the popular VGG13 as feature extractor for both the RGB input image $\mathcal{I}_l$ as well as the correspondent noisy disparity map $\mathcal{D}$. Notice that the two feature extractors do not share weights. The MLPs in charge of estimating the final disparity map, $\widetilde{\mathcal{D}}$, are implemented similarly to \cite{Saito2019ICCV}, though we replace the ReLU activations of the hidden layers by Sine functions \cite{sitzmann2019siren}. 

\textbf{Balanced Setup.} In the balanced stereo setting, $I_l$ and $I_r$ are at the same spatial resolution, thus the stereo blackbox computes an initial disparity map $\mathcal{D}$ at that same image size. With reference to Figure \ref{fig:network},  in such a scenario  $\mathcal{D}_\downarrow$ = $\mathcal{D}$ and $\mathcal{I}_l = \mathcal{I}_{l\downarrow}$, due to no downsampling/upsampling operations being  needed. We trained our refinement architecture, counting approximately 22.9 millions of parameters,  on the SceneFlow dataset for $100$ epochs setting the batch size to 8, using  random crops of $384 \times 384$ size as input to the network and sampling randomly and uniformly $N=30,000$ continuous 2D locations from each crop. We employed a learning rate of $10^{-4}$, halved after $80$ epochs. During training, we randomly alternate as stereo blackbox two popular traditional stereo algorithms, namely SGM \cite{hirschmuller2007stereo} and AD-Census \cite{Secaucus_1994_ECCV}. Moreover, we also provided a input corrupted \gt{} disparities obtained by adding different randomly-generated nuisances, like, e.g., Gaussian noise. By doing so, we force the network to  handle a variety of different noisy patterns. We set the maximum disparity value to $256$. A strong data augmentation procedure is applied to the RGB image as well as to the input disparity map, both normalized in the $[0,1]$ interval. In particular, the input disparities -and, accordingly, the ground-truths - are scaled by a factor randomly drawn from $[0.2,3]$. The training process takes approximately 24 hours. Further details are reported in the supplementary material.

\textbf{Unbalanced Setup.} When $\kappa \neq 1$ we tackle the unbalaced setting, where the two images of a stereo pair are captured at different image resolutions. In particular, without loss of generality,  we assume  $\kappa > 1$, that is $\mathcal{I}_l$ having a resolution higher than $\mathcal{I}_r$. We experiment with  stereo black-boxes realized by  existing methods, both traditional as well as based on deep networks. As such methods process two images at the same size in order to produce a disparity map, we first downsample the reference image $\mathcal{I}_l$ to match the lower resolution image $\mathcal{I}_r$ using bilinear interpolation, then  na\"ively upsample the computed disparity map $\mathcal{D_\downarrow}$ by  nearest neighbor interpolation so as to match the resolution  of $\mathcal{I}_l$ and obtain the actual map, $\mathcal{D}$, fed as input to our network. 
In our experiments, we use the UnrealStereo4K dataset for training following the protocol already described for the balanced setup but for the number of epochs and the crop size, set to 200 and $768 \times 768$, respectively.
As our architecture is not specifically designed to process very high resolution images, such as those provided by the UnrealStereo4K dataset  (8 Mpx), in order to train and validate our model we resize the reference image, $\mathcal{I}_l$, to $1920 \times 1080$ resolution. Thus, in the case of UnrealStereo4K, we receive a full-res $\mathcal{I}_l$ image and a low-res $\mathcal{I}_r$ (according to a certain unbalance factor $\kappa$) and then feed  $\phi_{\mathcal{I}}$ with a half-res $\mathcal{I}_l$, thereby constraining also the resolution of $\mathcal{D}$ to be $1920\times1080$. Nonetheless, to assess the ability of our architecture  to handle very high resolution stereo pairs, at test time we evaluate the performance by comparing the predicted disparities to the \gt{} available for the original full-res (8 Mpx) stereo pair. In fact, seamlessly evaluating at whatever output resolution regardless of the size(s) of the input images is a key trait of our architecture enabled by the proposed continuous formulation. 
In the experiments, we will compare with PSMNet \cite{chang2018psmnet} and HSMNet \cite{yang2019hierarchical}, trained respectively for 60 and 200 epochs with batches of 4 and 12 samples and the same crop size used for our model.

\textbf{Evaluation Metrics.} For evaluation, we adopt the standard end-point-error (EPE) metric obtained by averaging the absolute difference between the estimated disparity and the \gt{} as well as the percentage of pixels  having error higher than $th$ pixels, referred to in literature as \textit{bad-th}. Furthermore, we adopt the Soft-Edge-Error (SEE) metric, defined in \cite{chen2019over} to evaluate the correctness of the predicted disparities at object-boundaries and, thus, to assess the capability of a method to produce sharp depth discontinuities. In our experiments, we set the size of \gt{} patches to $5 \times 5$ when evaluating the SEE metric.

\subsection{Ablation study}
\label{sec:ablation}

In this section, we examine the importance of the proposed loss function and demonstrate the robustness of our network to diverse sources of input disparities. 

\textbf{Loss Function.} Table \ref{ablation} reports the performance of our architecture trained by  different loss functions. Specifically, following the balanced setup protocol described in \ref{sec:implementation_details}, we train our architecture using a na\"ive disparity regression $L_1$ loss, the mixture bimodal loss proposed in \cite{Tosi2021CVPR} to handle sharp disparity discontinuities and our proposed loss function (Eq. \ref{eq:loss}).  Notice that, for both the $L_1$ and the bimodal mixture output representations, a single MLP is in charge of regressing either the final disparity, for the former, or the five parameters of a univariate bimodal mixture distribution for the latter. In our architecture, we achieve this by modifying the last layer of  $MLP_C$ and dismissing  $MLP_O$. Then, we evaluate the trained models on the test set of the SceneFlow dataset  using  AD-Census \cite{Secaucus_1994_ECCV}, C-CNN \cite{luo2016efficient} and SGM \cite{hirschmuller2007stereo} as stereo black-boxes providing the map to be refined, so as assess upon the robustness of our method to diverse sources of input disparities. Firstly, we can observe how all the trained models improve the input disparity $\mathcal{D}$ by a large margin regardless the adopted loss function, proving the effectiveness of our architecture on the disparity refinement task with all the considered stereo black-boxes, and even when considering a method never seen at training time, such as C-CNN \cite{luo2016efficient}. Among the considered losses, the standard disparity regression produces worst results. Moreover, it is worth noticing that, although the bimodal mixture formulation and the proposed loss achieve rather similar results in terms of $SEE$, our output representation is consistently more accurate with respect to all the other error metrics, which vouches for the overall superiority of our proposal. 

\begin{table}[t]
\renewcommand{\tabcolsep}{10pt}
\centering
\scalebox{0.65}{
\begin{tabular}{c|c|rrrrrr}
     \hline
     Input & Method & \cellcolor{lower}bad2 & \cellcolor{lower}bad3 & \cellcolor{lower}bad4 & \cellcolor{lower}bad5 & \cellcolor{lower}EPE & \cellcolor{lower}SEE  \\
     \hline
     \multirow{4}{*}{\rotatebox[origin=c]{90}{AD-Census}} & $\mathcal{D}$ & 46.23 & 45.79 & 45.46 & 45.20 & 24.68 & 21.78 \\
     & $L_1$ & 13.94 & 9.81 & 7.57 & 6.16 & 1.86 & 3.14 \\
     & Bimodal & 9.37 & 6.54 & 5.11 & 4.25 & 1.66 & \textbf{1.47} \\
     & \cellcolor{ours} Ours & \cellcolor{ours} \textbf{8.49} & \cellcolor{ours} \textbf{6.10} & \cellcolor{ours} \textbf{4.86} & \cellcolor{ours} \textbf{4.10} & \cellcolor{ours} \textbf{1.53} & \cellcolor{ours} 1.48 \\
     \hline
     \multirow{4}{*}{\rotatebox[origin=c]{90}{C-CNN }} & $\mathcal{D}$ & 27.09 & 24.86 & 23.67 & 22.86 & 13.46 & 10.84 \\
     & $L_1$ & 10.54 & 7.69 & 6.14 & 5.14 & 1.68 & 2.86 \\
     & Bimodal & 7.93 & 5.66 & 4.56 & 3.90 & 1.64 & 1.38 \\
     & \cellcolor{ours} Ours & \cellcolor{ours} \textbf{7.11} & \cellcolor{ours} \textbf{5.25} & \cellcolor{ours} \textbf{4.28} & \cellcolor{ours} \textbf{3.68} & \cellcolor{ours} \textbf{1.42} & \cellcolor{ours} \textbf{1.36} \\
     \hline
      \multirow{4}{*}{\rotatebox[origin=c]{90}{SGM }} & $\mathcal{D}$ & 23.34 & 21.49 & 20.52 & 19.88 & 9.51 & 8.51 \\
      & $L_1$ & 8.86 & 6.39 & 5.13 & 4.32 & 1.37 & 5.63 \\
      & Bimodal & 6.08 & 4.49 & 3.64 & 3.11 & 1.25 & \textbf{1.17} \\
      & \cellcolor{ours} Ours & \cellcolor{ours} \textbf{5.80} & \cellcolor{ours} \textbf{4.33} & \cellcolor{ours} \textbf{3.56} & \cellcolor{ours} \textbf{3.07} & \cellcolor{ours} \textbf{1.19} & \cellcolor{ours} 1.26 \\
     \hline 
\end{tabular}
}

\caption{\textbf{Comparison between losses} on the SceneFlow test set. The task concerns refining the initial disparity map provided by different  blackboxes, \ie both handcrafted  (AD-Census\cite{Secaucus_1994_ECCV}, SGM\cite{hirschmuller2007stereo} ) and learned (C-CNN\cite{luo2016efficient}) stereo matchers. We report the results obtained by our network when trained by a standard $L_1$, a bimodal mixture representation \cite{Tosi2021CVPR} and our proposed loss.}
\label{ablation}
\end{table}

\textbf{End-to-End Stereo Blackbox.}
We also investigate the capability of our architecture to refine the disparity maps predicted by state-of-the-art deep stereo networks. In particular, in Table \ref{tab:end_to_end_refinement} we consider several deep architectures trained on SceneFlow as stereo black-boxes  and evaluate the refined disparities yielded by our method on the 387 SceneFlow test images. Consistently to  Table \ref{ablation}, our method successfully ameliorates the initial disparity maps for all the considered stereo blackboxes  on both the \textit{bad3} and the SEE metrics, thus proving that our architecture is also be beneficial when deployed in conjunction with deep stereo models. Again, it is worth highlighting that none of the networks considered in Tab. \ref{tab:end_to_end_refinement} was used as stereo blackbox to train our architecture. Yet, it can be observed how the EPE score slightly increase in case of the highly accurate disparity maps computed by top-performing stereo networks, such as GANet and AANet. We regard this as a trade-off associated with a loss aimed at better capturing depth edges, like ours (Eq. \ref{eq:loss}), compared to a standard regression loss. In fact, stereo networks trained by standard regression losses produce over-smoothed disparities at object-boundaries that are not penalized by the EPE metric but lead to severe bleeding artifacts when converted to 3D point clouds. Conversely, our approach takes sharp predictions at edges which, when wrong, may cause larger errors and slightly higher EPEs. However, sharp depth discontinuities result in clear and more realistic  3D point clouds, as shown qualitatively in the supplementary material. 

\begin{table}[t]
\centering
\scalebox{0.6}{
\setlength{\tabcolsep}{22pt}
\begin{tabular}{l|rrrrrr}
\hline
 Input & \cellcolor{lower}bad3 & \cellcolor{lower}SEE & \cellcolor{lower}EPE \\
\hline
GANet \cite{zhang2019ga} & 3.55 &  1.83 &  \textbf{0.95}\\
\rowcolor{ours}
GANet \cite{zhang2019ga}+ Ours & \textbf{3.44} & \textbf{1.43}  & 0.96 \\
\hline
AANet \cite{xu2020aanet} & 4.12 & 2.81 & \textbf{1.10} \\
\rowcolor{ours}
AANet \cite{xu2020aanet} + Ours & \textbf{3.81} & \textbf{1.62} & 1.14 \\
\hline
HSMNet \cite{yang2019hierarchical} & 8.02 & 3.77 & 1.86 \\
\rowcolor{ours}
HSMNet \cite{yang2019hierarchical} + Ours & \textbf{6.13} & \textbf{1.86} & \textbf{1.53}\\
\hline
PSMNet \cite{chang2018psmnet} & 7.98 & 2.96 & 1.87 \\
\rowcolor{ours}
PSMNet \cite{chang2018psmnet} + Ours & \textbf{6.94} & \textbf{1.85} & \textbf{1.71}\\
\hline
\end{tabular}
}
\caption{\textbf{End-to-End Networks as Stereo Blackbox.}  We validate our model using disparity maps computed by several end-to-end stereo network.  For all the networks, we used the official weights released by the authors after  training on SceneFlow.}
\label{tab:end_to_end_refinement}
\end{table}

\subsection{Balanced Setup}

In this section, we conduct experiments considering the standard balanced stereo setup. In particular,  first we compare the proposed architecture to the main existing deep networks designed to pursue disparity refinement. Then, we assess the capability to generalize to unseen data comparatively with respect to DDR  \cite{gidaris2017detect}, \ie the only refinement network that has been evaluated also on the Middlebury v3 dataset. Finally, we compare our proposal to several stereo methods and across different real-world scenarios. This highlights how, by deploying a traditional stereo matcher \cite{hirschmuller2007stereo} as blackbox, our architecture yields superior zero-shot generalization even with respect to the recent end-to-end networks specifically designed to achieve this capability.

\subsubsection{Comparison to Existing Refinement Methods}

\textbf{Comparison to Refinement Strategies.} We compare our proposal to the state-of-the-art published methods on the online KITTI 2015 leaderboard. In order to be compliant with the competitors, similarly to \cite{batsos2018recresnet, gidaris2017detect, Jie_2018_CVPR}, we started from the model pre-trained on SceneFlow and then fine-tuned it by the 200 images of the KITTI 2015 training set based on the available sparse \gt{} disparities.  The first part of Table \ref{KITTI_2015_online_benchmark} reports the results of our submission alongside those of other competing refinement methods: our architecture achieves state-of-the-art results in all metrics (D1-all, D1-fg and D1-bg), clearly outperforming all the other refinement techniques. In the second  part of Table \ref{KITTI_2015_online_benchmark} we also report the results achieved on KITTI 2015 by several end-to-end stereo networks: it is worth highlighting how our refinement architecture yields \emph{in-domain} performance comparable with respect to these latter. 

\begin{table}[t]
\setlength{\tabcolsep}{30pt}
\centering
\scalebox{0.55}{
\begin{tabular}{l|c|c|c}
\hline
Method & \cellcolor{lower}\textbf{D1-all} & \cellcolor{lower}D1-fg & \cellcolor{lower}D1-bg \\
\hline
\multicolumn{4}{c}{Disparity Refinement} \\
\hline
Dil-Net \cite{ferrera2019fast} & 3.92 & 7.44 & 3.22  \\
DRR $\times 2$  \cite{gidaris2017detect} \textdagger & 3.16 & 6.04 & 2.58  \\
LRCR \cite{Jie_2018_CVPR} &  3.03 & 5.42 & 2.55\\
RecResNet \cite{batsos2018recresnet} & 3.10 & 6.30 & 2.46  \\
\rowcolor{ours}
Ours \textdagger & \textbf{2.35} & \textbf{3.93} & \textbf{2.03}   \\
\hline
\multicolumn{4}{c}{End-to-End Stereo} \\
\hline
AANet \cite{xu2020aanet} & 2.55 & 5.39 & 1.99  \\
PSMNet \cite{chang2018psmnet} & 2.32 & 4.62 & 1.86  \\
HITNet \cite{Tankovich_2021_CVPR} & 1.98 & 3.20 & 1.74   \\
DSMNet \cite{zhang2019domaininvariant} & \textbf{1.77} & \textbf{3.23} & \textbf{1.48} \\
\hline
\end{tabular}
}
\caption{\textbf{Evaluation on KITTI 2015 Benchmark.} Methods indicated with \textdagger{} consider  the disparity maps computed by C-CNN \cite{luo2016efficient} as noisy input of the network. The other refinement methods adopt different noisy disparity inputs as described in the papers.}
\label{KITTI_2015_online_benchmark}
\end{table}

{
\textbf{Comparison with refinement frameworks}. We compare our method to other refinement models \cite{gidaris2017detect,batsos2018recresnet,ferrera2019fast}. Table \ref{DRR_comparison} reports the results obtained on Middlebury v3 (a) and KITTI 2015 (b), (c). In the former case, we compare with DRR \cite{gidaris2017detect} after training on the SceneFlow dataset. For the sake of evaluation, the two methods adopt the same noisy input disparities computed by a deep patch matching approach \cite{luo2016efficient} trained on KITTI. Notice that, differently from \cite{gidaris2017detect}, at training time our model is fed only with noisy inputs extracted by traditional stereo matchers, \ie  SGM and AD-Census. Despite this, our method notably outperforms DRR by a large margin in all metrics. 

In (b) and (c) we collect results on KITTI 2015, for which we fine-tune our model on the first 160 images of the training set and evaluate on the remaining 40, the same setting followed by DRR \cite{gidaris2017detect}, RecResNet \cite{batsos2018recresnet} and Dil-Net \cite{ferrera2019fast}. Again, our model at training time is fed only with inputs obtained through SGM and AD-Census, while at testing time it refines disparity maps by C-CNN \cite{luo2016efficient} (b) or OpenCV SGBM implementation (c), outperforming DRR and RecResNet in the former case, Dil-Net in the latter.
}

\begin{table}[t]
\centering
\renewcommand{\tabcolsep}{2pt}
\scalebox{0.48}{
\begin{tabular}{ccccc}
\begin{tabular}{l|cc|cc}
\multicolumn{1}{c}{ } & \multicolumn{4}{c}{Middlebury v3} \\
\hline
 & \multicolumn{2}{c|}{\cellcolor{lower}bad2} &  \multicolumn{2}{c}{\cellcolor{lower}EPE} \\
\hline
Method & Non-Occ & All & Non-Occ & All \\
\hline
C-CNN \cite{luo2016efficient} & 18.24 & 26.71 & 6.06 &  8.71 \\
\hline
DRR \cite{gidaris2017detect} & 12.85 & 17.83  & 1.77 & 2.37 \\
DRR $\times 2$  \cite{gidaris2017detect} & 11.53 & 16.41 & 1.79 & 2.32 \\
\rowcolor{ours}
Ours & \textbf{10.84} & \textbf{15.02} & \textbf{1.38} & \textbf{1.84}  \\
\hline
\rowcolor{white}
\multicolumn{5}{c}{\textbf{(a)}}\\
\end{tabular}
& &
\begin{tabular}{l|cc|cc}
\multicolumn{1}{c}{ } & \multicolumn{4}{c}{KITTI 2015} \\
\hline
 & \multicolumn{2}{c|}{\cellcolor{lower}bad3} &  \multicolumn{2}{c}{\cellcolor{lower}EPE} \\
\hline
Method & Non-Occ & All & Non-Occ & All \\
\hline
C-CNN \cite{luo2016efficient} & 6.41 & 8.25 & 1.70 & 2.46 \\
\hline
DRR $\times 2$  \cite{gidaris2017detect} & 2.58 & 3.08 & 0.78 & 0.84\\
RecResNet \cite{batsos2018recresnet} & - & 3.46 & - & - \\
\rowcolor{ours}
Ours & \textbf{2.27} & \textbf{2.60} & \textbf{0.75} & \textbf{0.79}  \\
\hline
\rowcolor{white}
\multicolumn{5}{c}{\textbf{(b)}}\\
\end{tabular}
& &
\begin{tabular}{l|c}
\multicolumn{2}{c}{KITTI 2015} \\
\hline
 & {\cellcolor{lower}bad3}\\
\hline
Method & All \\
\hline
SGBM \cite{hirschmuller2007stereo} & 5.15 \\
\hline
Dil-Net (ref.) \cite{ferrera2019fast} & 4.58 \\
Dil-Net (fus.) \cite{ferrera2019fast} & 3.07 \\
\rowcolor{ours}
Ours & \textbf{2.93}\\
\hline
\rowcolor{white}
\multicolumn{2}{c}{\textbf{(c)}}\\
\end{tabular}

\end{tabular}
}
\caption{\textbf{Comparison with refinement frameworks}. In (a), all models are trained on SceneFlow dataset and tested on the 15 images of the Middlebury v3 training dataset at quarter resolution. In (b) and (c), models are fine-tuned on the first 160 images of KITTI 2015 training set and evaluated on the remaining 40.}
\label{DRR_comparison}
\end{table}

\subsubsection{Zero-Shot Generalization}\label{sec:zero-shot}

Finally, we evaluate the generalization ability of our  method using three different real-world datasets (KITTI, Middlebury v3 and ETH3D). We compare our network architecture to traditional stereo techniques as well as to recent state-of-the-art end-to-end stereo networks. For fairness, all networks are trained in a supervised setting on the SceneFlow synthetic dataset only.  Table \ref{synthetic-to-real} shows how our architecture, when fed with input disparity maps computed by a traditional stereo algorithm such as SGM \cite{hirschmuller2007stereo}, consistently achieves the highest accuracy on all the considered datasets. It is particularly remarkable how our framework outperforms even the end-to-end stereo networks specifically designed for robust cross domain generalization \cite{zhang2019domaininvariant,cai2020matchingspace}.

\begin{table}[t]
\centering
\scalebox{0.65}{
\renewcommand{\tabcolsep}{10pt}
\begin{tabular}{l|rr|rrr|r}
\hline
& \multicolumn{2}{c|}{\cellcolor{lower}bad3} &  \multicolumn{3}{c|}{\cellcolor{lower}bad2} & {\cellcolor{lower}bad1} \\
\hline
Target Domain & \multicolumn{2}{c|}{KITTI} & \multicolumn{3}{c|}{Middlebury v3} & ETH3D \\
 & 2012 & 2015 & Full & Half & Quarter &  \\
\hline
  
SGM \cite{hirschmuller2007stereo} & 14.7 & 14.0 &  27.6 & 23.2 & 17.7 & 13.4\\
\hline
PSMNet \cite{chang2018psmnet} & 27.8 & 30.7 & 39.5 & 25.1 & 14.2 & 23.8\\
GANet \cite{zhang2019ga} & 10.1 & 11.7 & 32.2 & 20.3 & 11.2 & 14.1\\
HITNet \cite{Tankovich_2021_CVPR} & 6.4 & 6.5 & - & - & - & -\\
MS-GCNet \cite{cai2020matchingspace} & *5.5 & 6.2 & - & 18.5 & - & 8.8\\
DSMNet \cite{zhang2019domaininvariant} & 6.2 & 6.5 & 21.8 & 13.8 & 8.1 & 6.2\\
\hline
\rowcolor{ours}
SGM \cite{hirschmuller2007stereo} + Ours & \textbf{6.0}/\textbf{*5.0} & \textbf{5.5} & \textbf{19.2}  & \textbf{12.4} & \textbf{7.9} & \textbf{4.8}\\
\hline
\end{tabular}
}
\caption{\textbf{Generalization Performance.} All methods are trained on SceneFlow and tested on the KITTI, Middlebury v3, and ETH3D datasets. Errors are the percentage of pixels with EPE greater than the specified threshold. We use the standard evaluation thresholds: 3px for KITTI, 2px for Middlebury v3, 1px for ETH3D. In KITTI the results labeled with * denote that occluded pixels have not been considered in the evaluation.}
\label{synthetic-to-real}
\end{table}

\begin{figure}
    \centering
    \renewcommand{\tabcolsep}{1pt}   
    \scalebox{0.72}{
    \begin{tabular}{ccc}
    
        \includegraphics[width=0.18\textwidth]{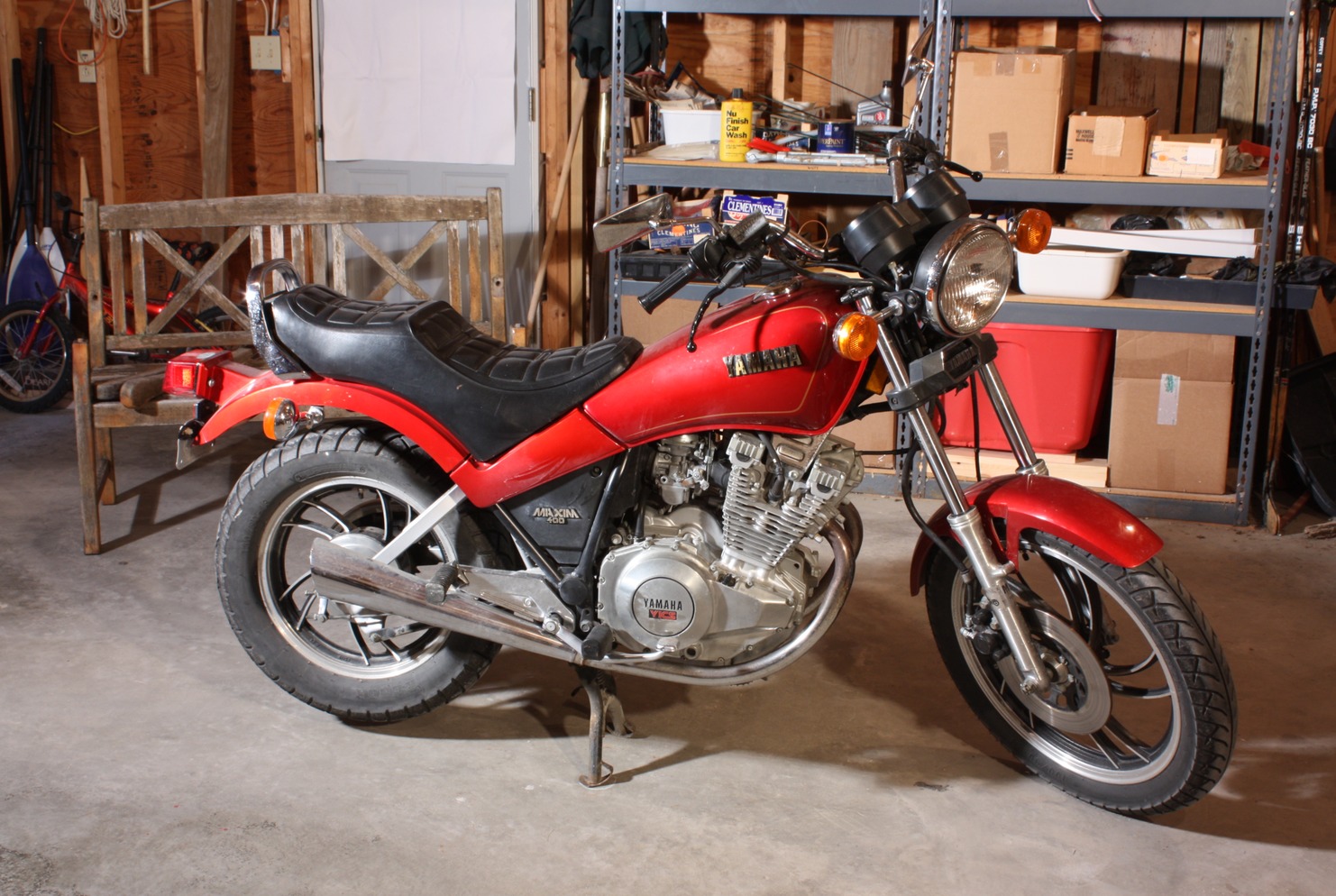}  &
        \includegraphics[width=0.18\textwidth]{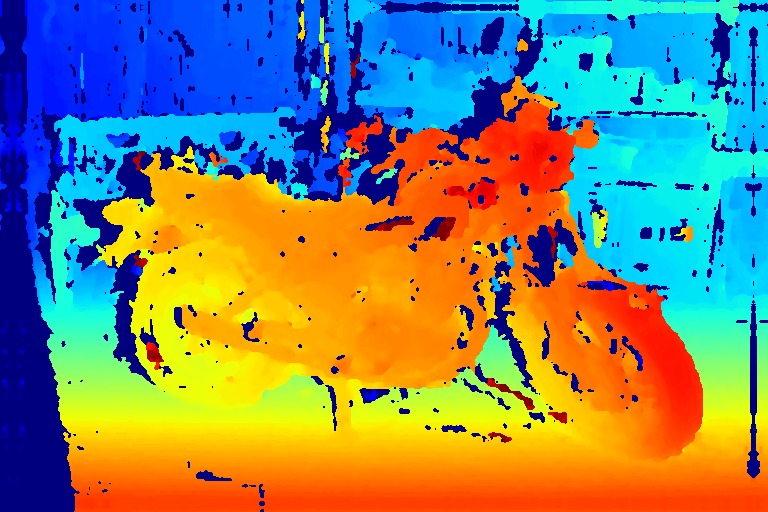}  &
        \includegraphics[width=0.18\textwidth]{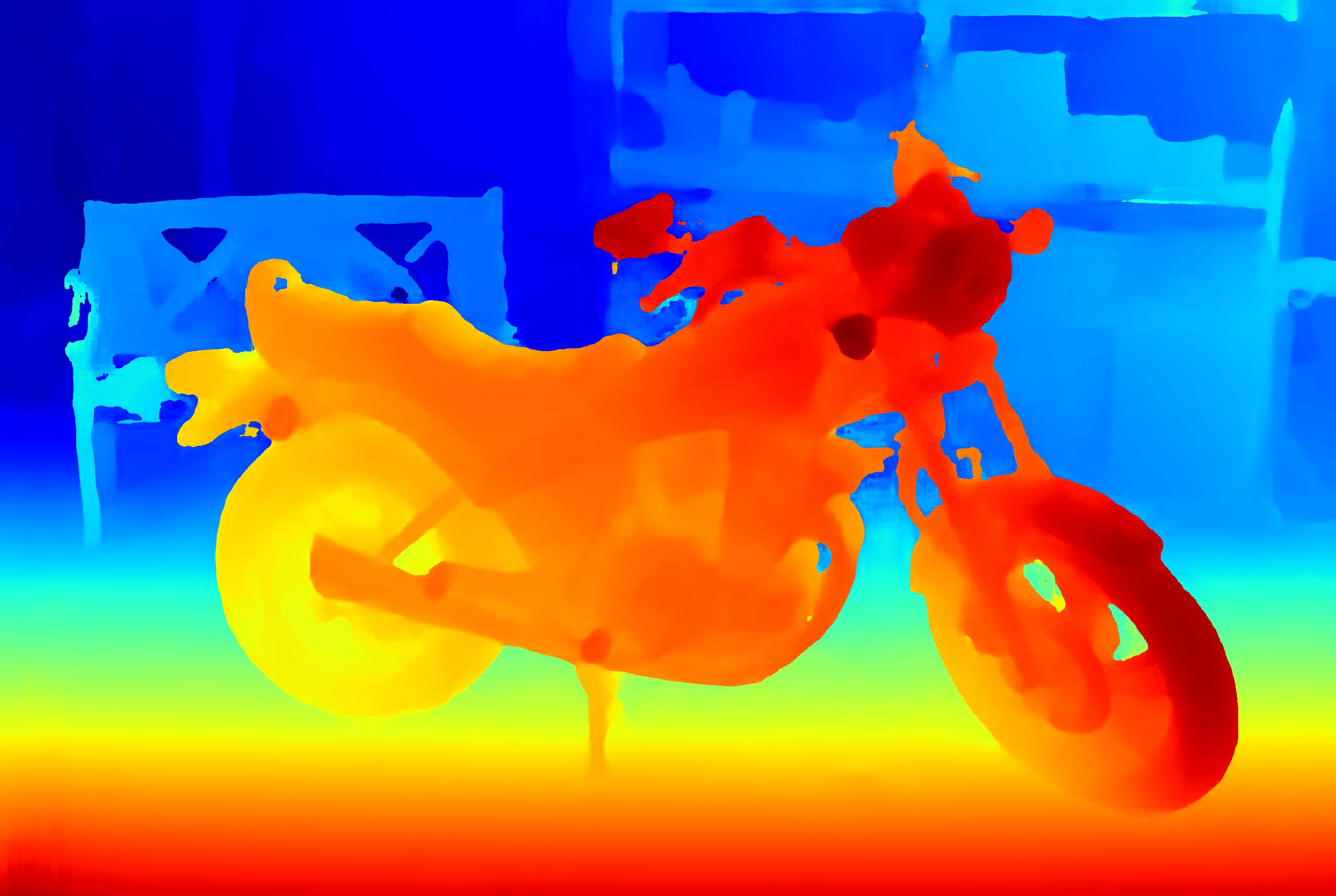} \\
        
        \includegraphics[width=0.18\textwidth]{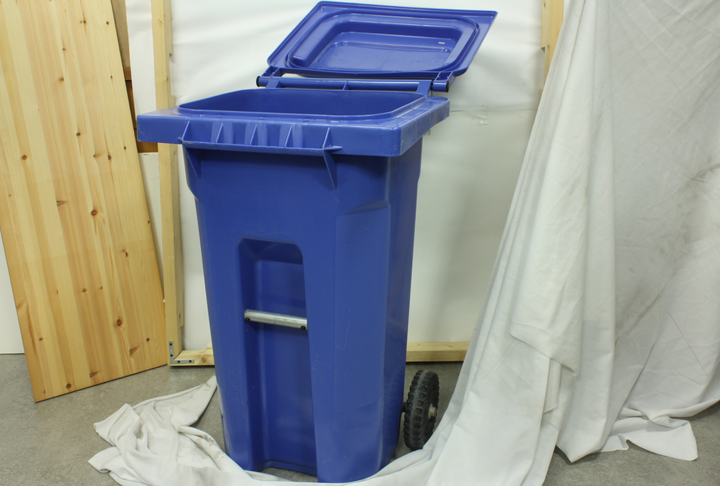}  &
        \includegraphics[width=0.18\textwidth]{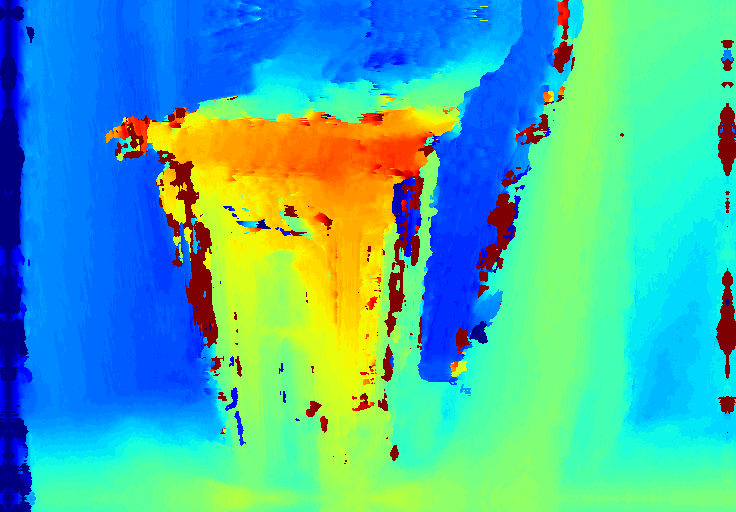}  &
        \includegraphics[width=0.18\textwidth]{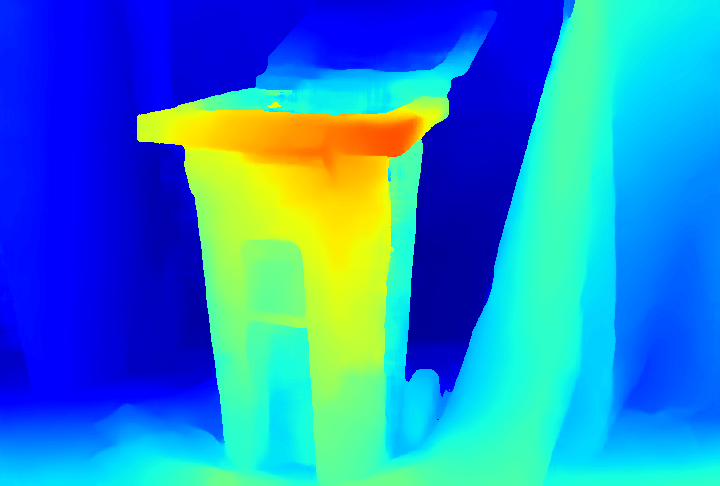} 
        \\

        $\mathcal{I}_l$ & $\mathcal{D}$ & $\mathcal{\widetilde{D}}$
        \\
    \end{tabular}
    }
    \caption{\textbf{Qualitative Results on Middlebury v3 - balanced setup}. From left to right, the input image $\mathcal{I}_l$ from Middlebury v3, the disparity map $\mathcal{D}$ produced by SGM (top) and C-CNN (bottom) and our refined disparity $\mathcal{\widetilde{D}}$.
    }
    \label{fig:qualitative_balanced}
\end{figure}

\subsection{Unbalanced Setup}

In this section, we demonstrate how our architecture can be effectively deployed to handle unbalanced stereo images. To this aim, we experiment with stereo images acquired at  different resolutions by adopting the synthetic high-resolution UnrealStereo4K dataset, that allows us to simulate different unbalance factors, $\kappa$, as well as to evaluate the accuracy of the estimated disparities using \gt{} information. Similarly to the balanced setup, we also assess out-of-domain generalization performance by testing on  Middlebury v3 without any fine-tuning.

\subsubsection{Handling Unbalanced Stereo Images} Table \ref{Unbalanced_MPI} reports our experimental study dealing with the unbalanced stereo setting on the UnrealStereo4K dataset.
We assume two baseline approaches to deal with this setting, namely  i) downsampling $\mathcal{I}_l$ to the same low-resolution as $\mathcal{I}_r$, estimating disparity and finally upsampling it by bilinear interpolation to the resolution of $\mathcal{I}_l$  or ii) upsampling the low-res $\mathcal{I}_r$ to the same resolution as $\mathcal{I}_l$ and directly estimating disparity at the resolution of $\mathcal{I}_l$.
We run experiments starting from  three stereo methods: SGM, PSMNet \cite{chang2018psmnet} and HSMNet \cite{yang2019hierarchical}. According to the considered  method, one approach is preferred to the other. Indeed, for SGM and PSMNet we adopt i), as both methods cannot process  high-res images due to memory constraints, while for HSMNet we adopt the latter, since its architecture is specifically designed to handle high-res images.
We train PSMNet and HSMNet on the official training split of the dataset, and evaluate them on the test set using $\kappa=4, 8$ and $12$. Again, we train a single instance of our neural refinement network and test its accuracy when refining the raw disparities provided by SGM, PSMNet and HSMNet.

\begin{table}[t]
\centering
\renewcommand{\tabcolsep}{7pt}
\scalebox{0.72}{
\begin{tabular}{l|cc|cc|cc}
\hline
&  \multicolumn{2}{c}{$\kappa=4$} & \multicolumn{2}{c}{$\kappa=8$} & \multicolumn{2}{c}{$\kappa=12$} \\
\hline
Method & \cellcolor{lower}bad3 & \cellcolor{lower}EPE & \cellcolor{lower}bad3 & \cellcolor{lower}EPE & \cellcolor{lower}bad3 & \cellcolor{lower}EPE \\
\hline
& \multicolumn{6}{c}{Traditional Stereo} \\
\hline
SGM \cite{hirschmuller2007stereo} & 26.33 & 41.56 & 37.74 & 43.27  & 50.69 & 45.45 \\
\rowcolor{ours}
SGM \cite{hirschmuller2007stereo} + Ours & \textbf{12.21} & \textbf{5.65} & \textbf{15.92} & \textbf{7.37} & \textbf{22.06} & \textbf{8.04} \\
\hline
& \multicolumn{6}{c}{End-to-End Stereo} \\
\hline
PSMNet \cite{chang2018psmnet} &  15.22 & 4.37 & 17.83 & 4.67 &  42.69 & 7.42 \\
\rowcolor{ours}
PSMNet \cite{chang2018psmnet} + Ours &  \textbf{12.45} & \textbf{3.86} & \textbf{14.17} & \textbf{4.06} & \textbf{35.98} & \textbf{6.22} \\
\hline
HSMNet \cite{yang2019hierarchical} & 15.31 & 6.73 & 29.07 & 9.25 & 43.14 & 13.40 \\
\rowcolor{ours}
HSMNet \cite{yang2019hierarchical} + Ours & \textbf{12.12} & \textbf{5.61} & \textbf{20.36} & \textbf{6.90} & \textbf{31.67} & \textbf{9.43} \\
\hline
\end{tabular}
}
\caption{\textbf{Experimental Study of Unbalanced Setups.} We rely on the UnrealStereo4K dataset to evaluate different methods in  unbalanced setups featuring different $\kappa$ factors.}
\label{Unbalanced_MPI}
\end{table}

\begin{figure}[t]
    \centering
    \renewcommand{\tabcolsep}{1pt}   
    \begin{tabular}{ccc}
        \includegraphics[width=0.15\textwidth]{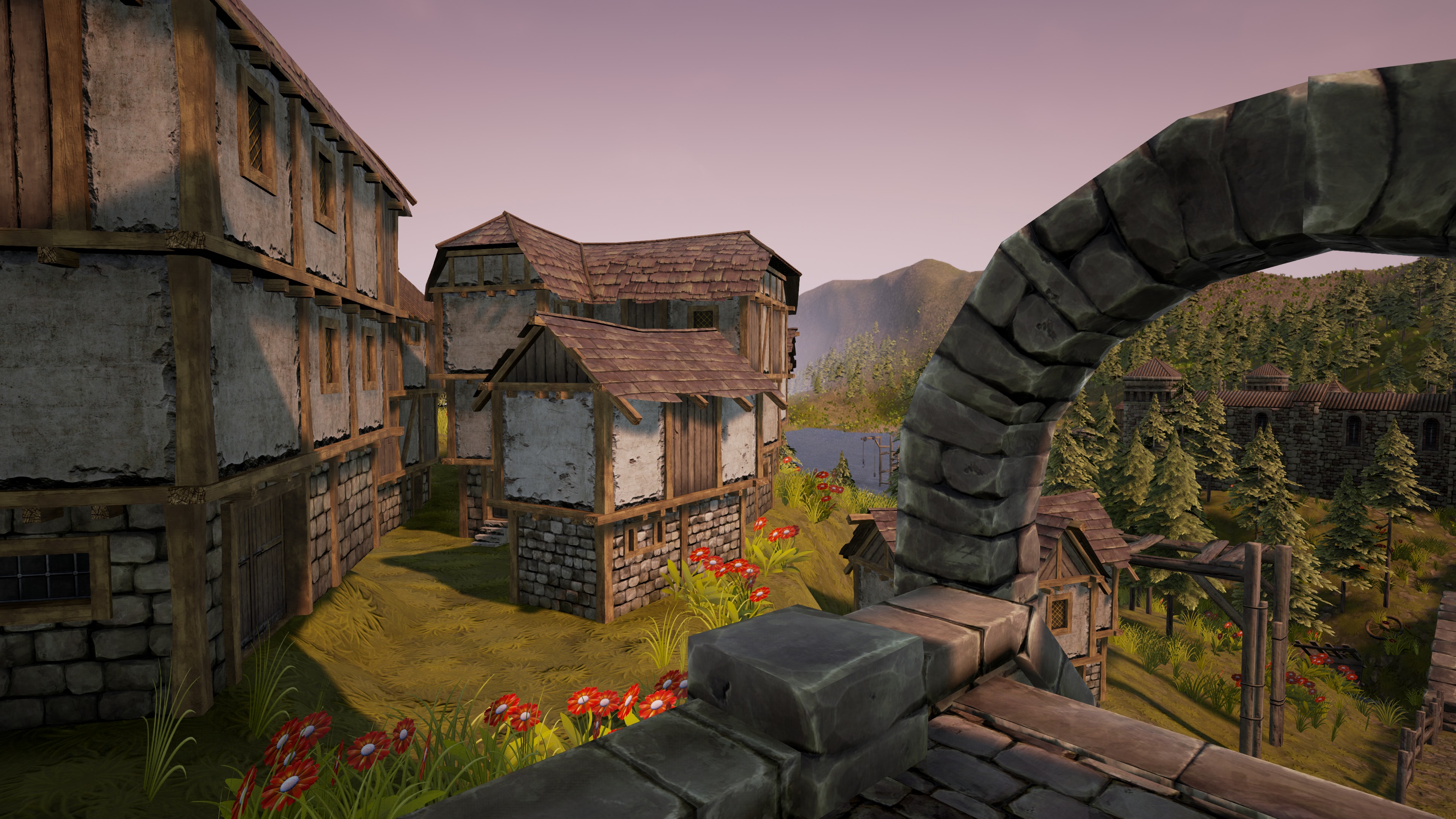}  &
        \includegraphics[width=0.15\textwidth]{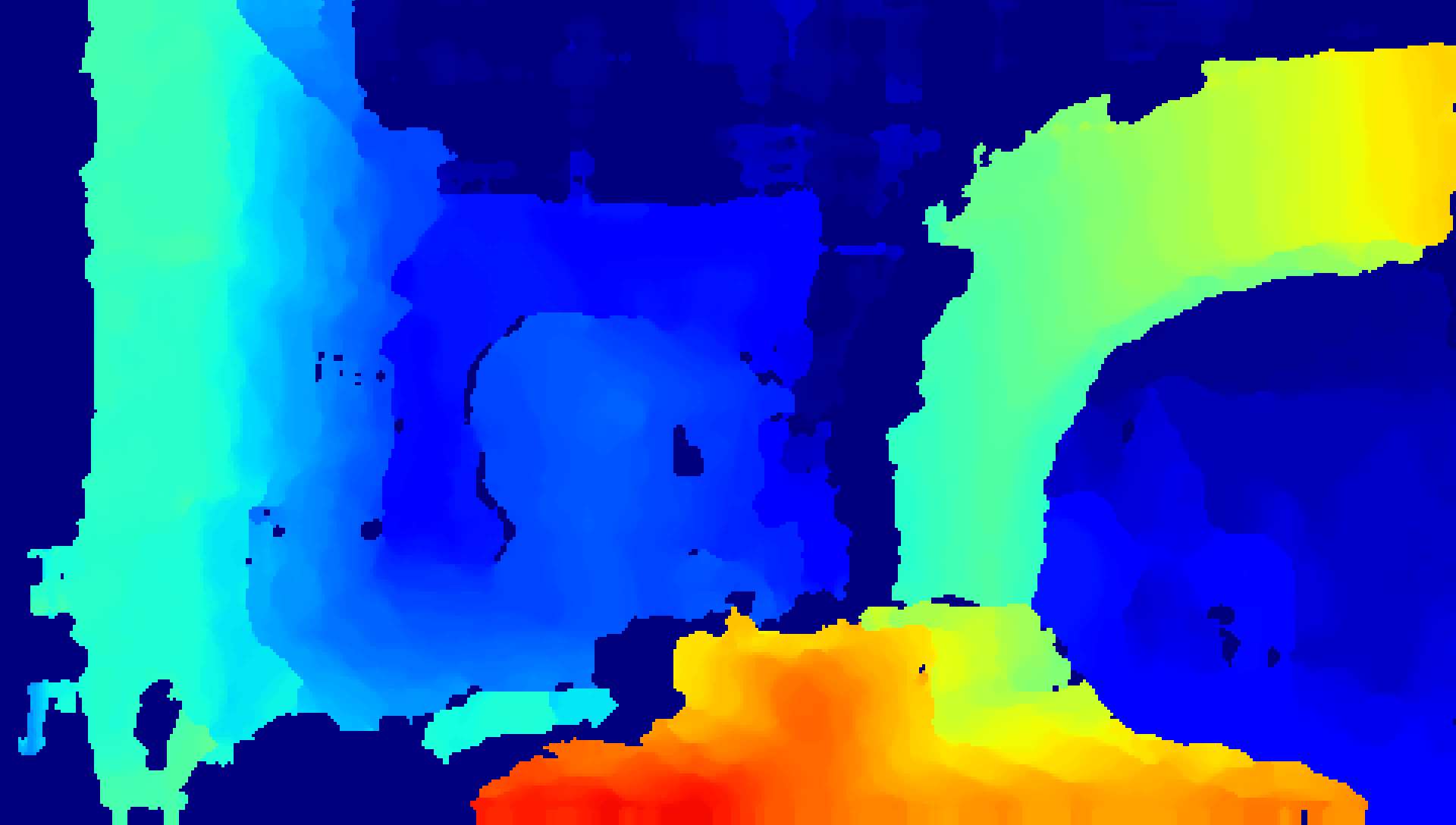} &
        \includegraphics[width=0.15\textwidth]{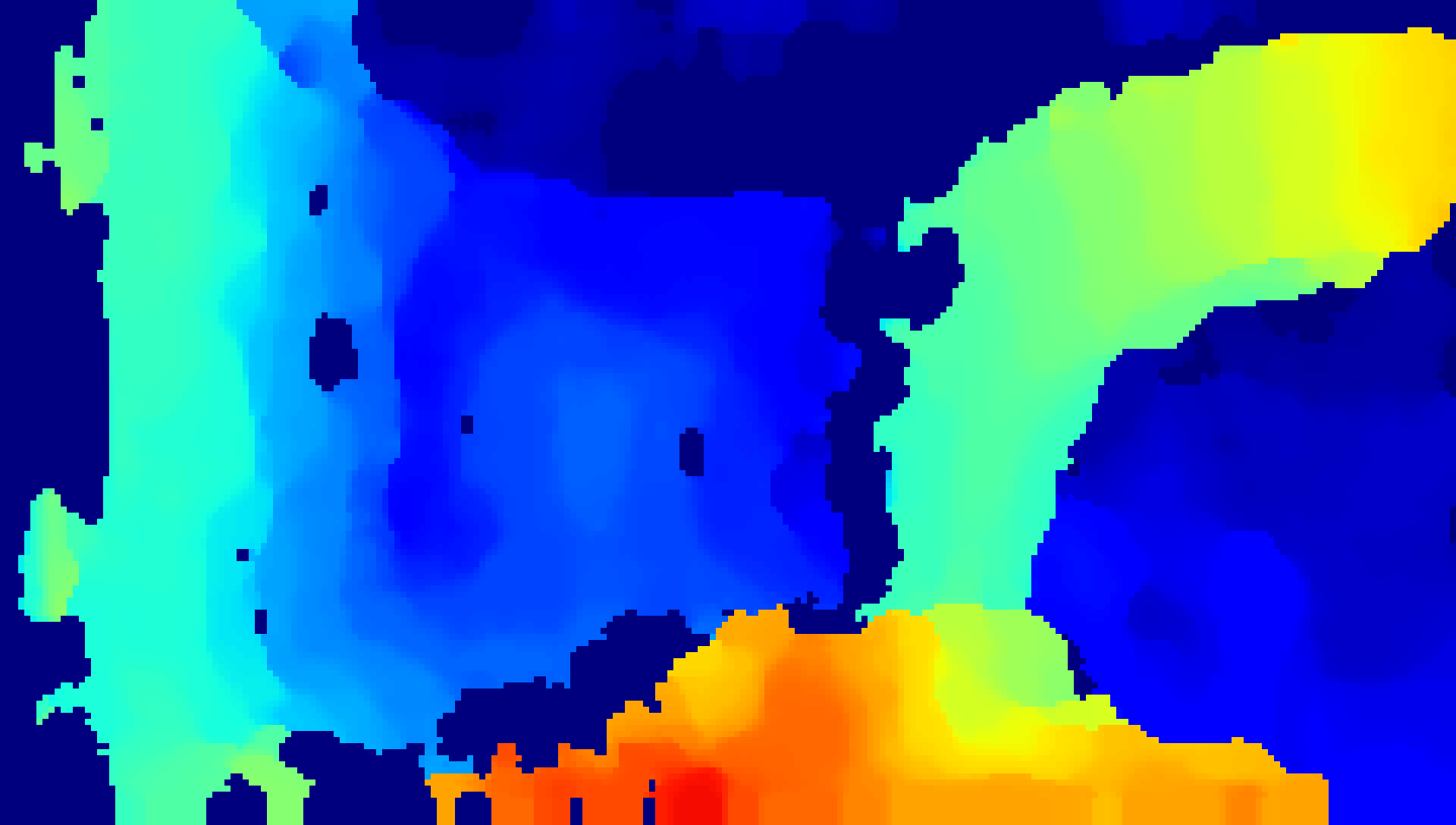} \\
        \scriptsize $\mathcal{I}_L$ & \scriptsize $\mathcal{D} (\kappa=8)$ & \scriptsize $\mathcal{D} (\kappa=12)$ \\
        \includegraphics[width=0.15\textwidth]{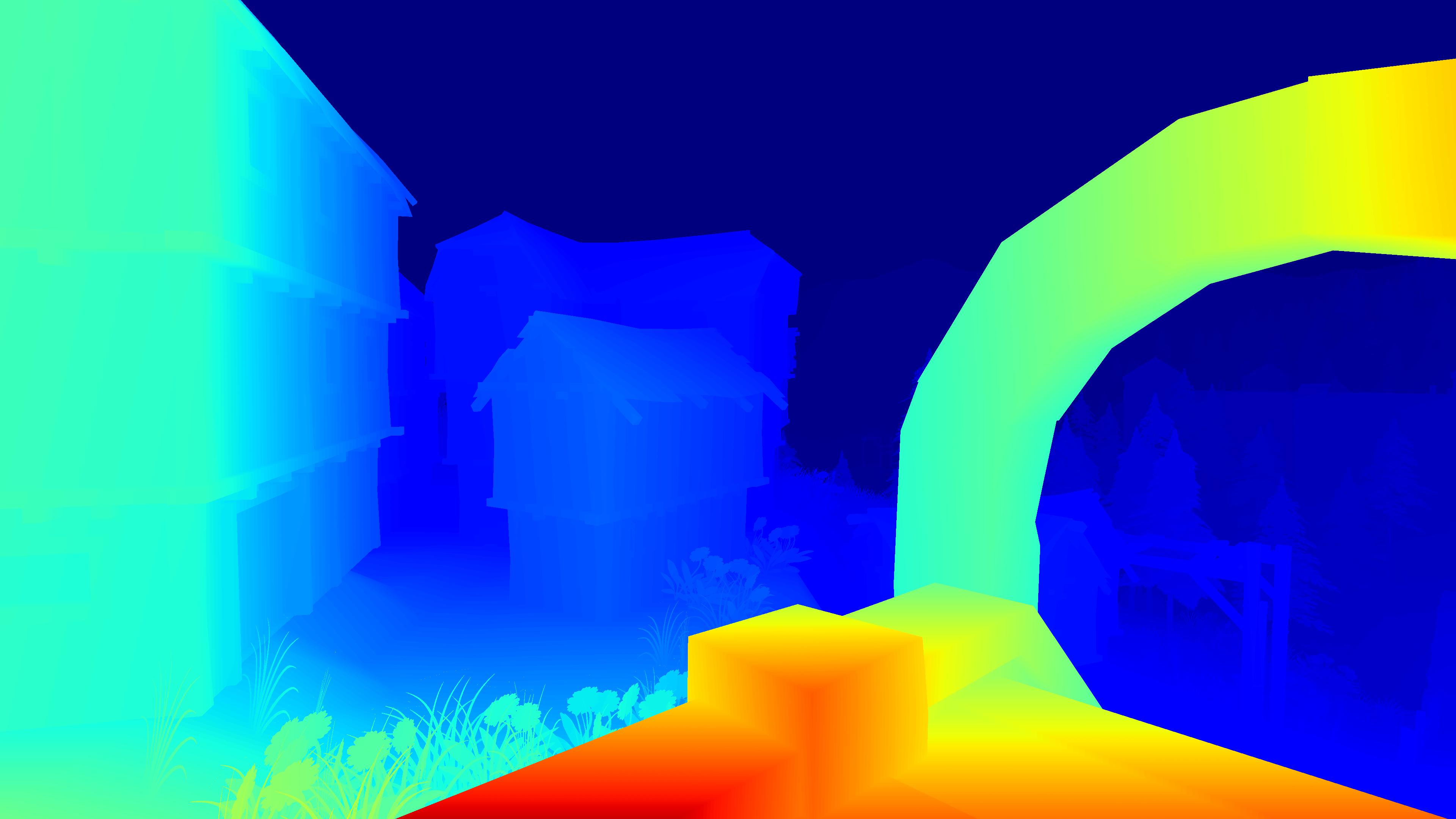}  & 
        \includegraphics[width=0.15\textwidth]{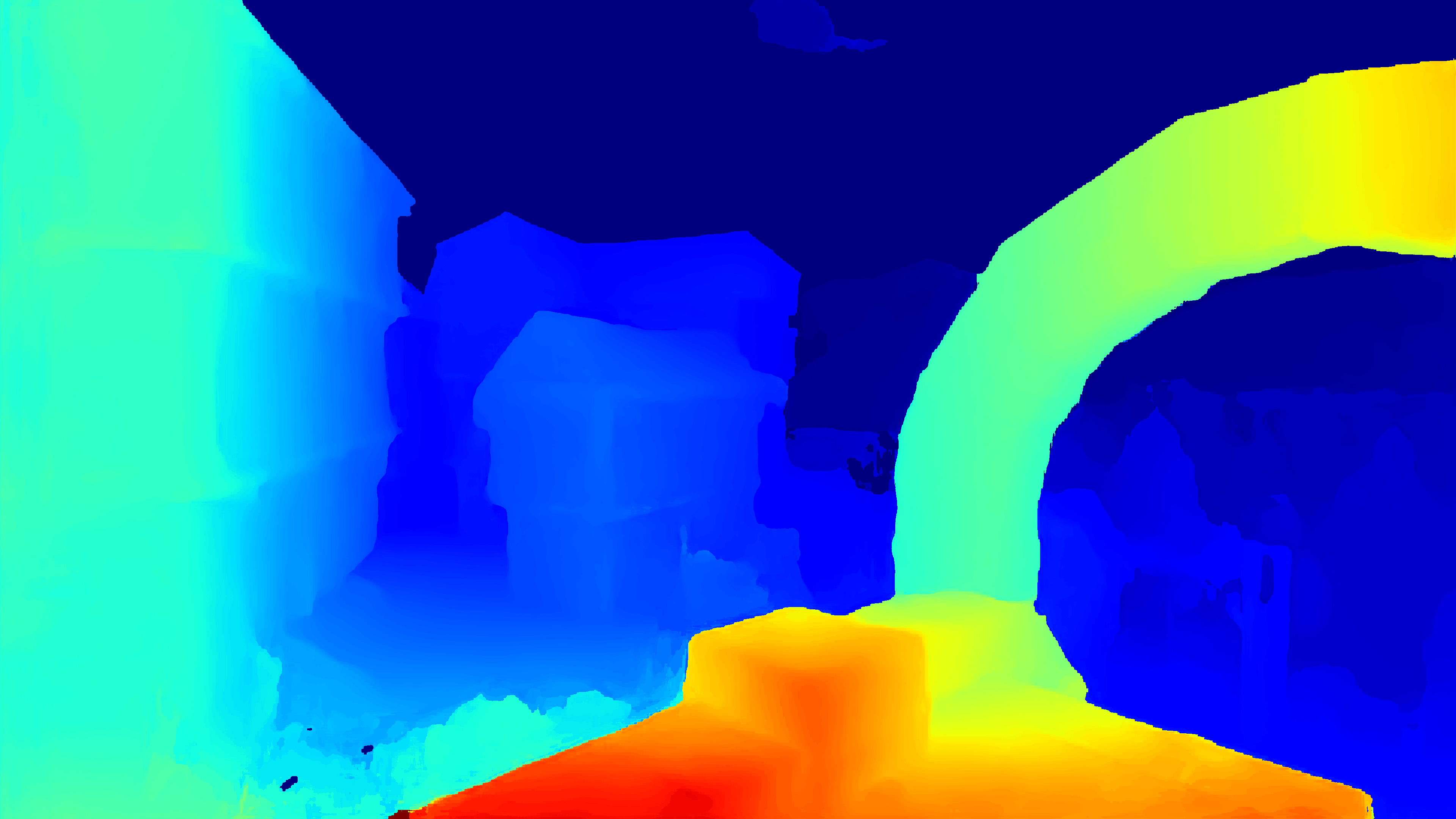} &
        \includegraphics[width=0.15\textwidth]{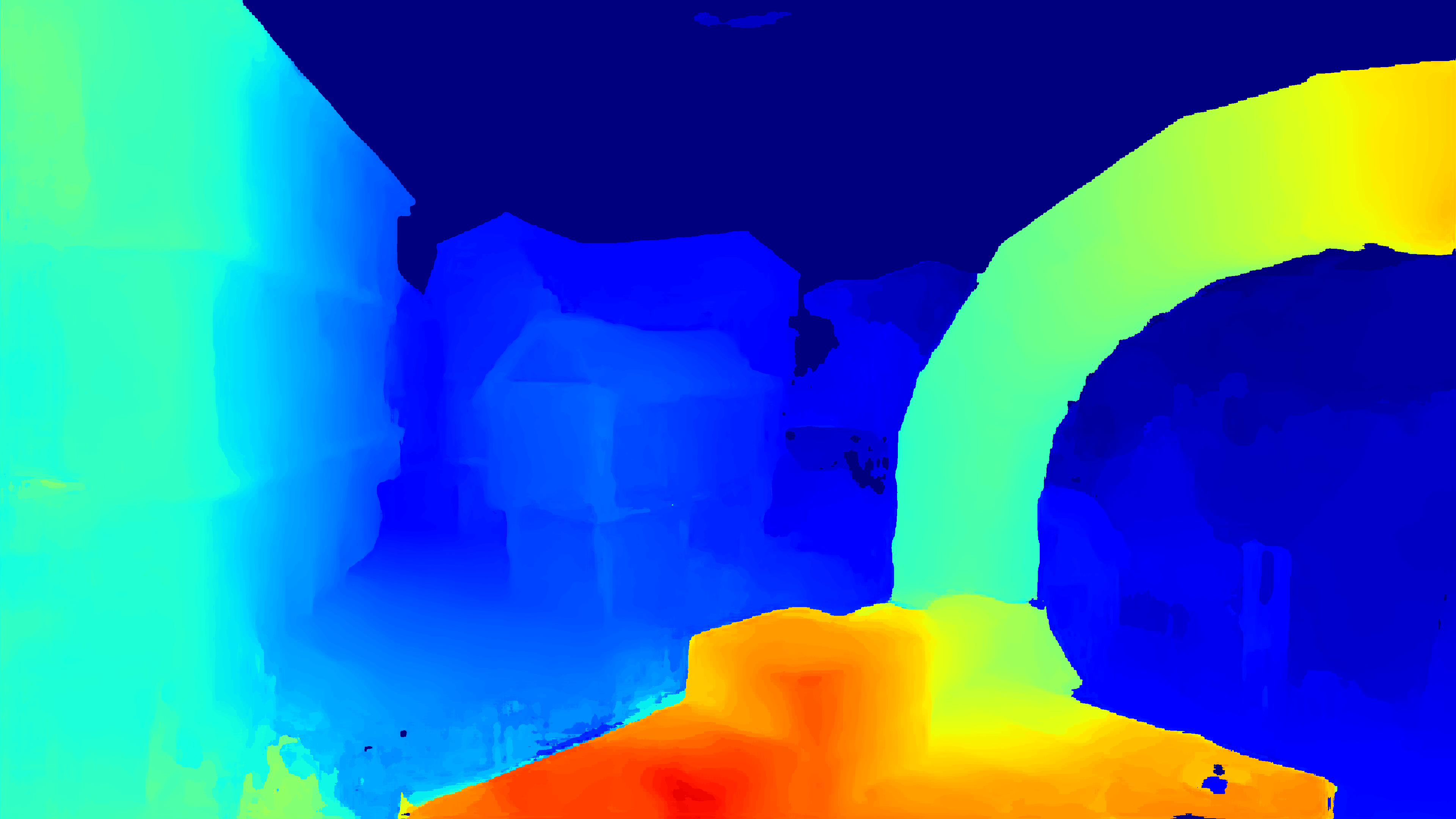} \\
        \scriptsize Ground-Truth & \scriptsize $\mathcal{\widetilde{D}} (\kappa=8)$ & \scriptsize $\mathcal{\widetilde{D}} (\kappa=12)$  \\
    \end{tabular}
    \caption{\textbf{Qualitative Results on UnrealStereo4K -- unbalanced setting.} The top row depicts the input image, $\mathcal{I}_l$, at $3840 \times 2160$ and the disparity maps, $\mathcal{D}$, computed by SGM when the right image, $\mathcal{I}_r$, is $480\times270$  and $320\times180$ ($\kappa=8$ and $12$). The bottom row shows \gt{} and estimated disparity $\mathcal{\widetilde{D}}$ at $3840 \times 2160$.}
    \label{fig:qualitative_unbalanced}
\end{figure}

We can observe how applying our refinement strategy yields consistently  a significant accuracy improvement with all the considered methods. By taking a deeper look we can also notice that applying our strategy to the three methods produces almost equivalent bad3  results in case of the lowest unbalance factor, \ie{} $\kappa=4$. In this case, HSMNet yields the best bad3 results. With a larger unbalance factor, as in the case of $\kappa=12$, the number of outliers increases significantly when refining disparity maps produced by deep networks, while processing the outcome of SGM leads to the best result, with only a 10\% bad3 increase with respect to the $\kappa=4$ setting, whereas PSMNet and HSMNet yield about +20\% bad3. Concerning EPE, refining the predictions by PSMNet consistently produces the lowest error.

Finally, we show in Figure \ref{fig:qualitative_unbalanced} how our method can produce sharp and detailed high-res $\mathcal{\widetilde{D}}$ disparity maps even in case of very unbalanced setups, such as $\kappa=12$.

\begin{table}[t]
    \centering
    \renewcommand{\tabcolsep}{18pt}
    \scalebox{0.65}{
    \begin{tabular}{c|c|cccc}
        & & \cellcolor{lower}bad2 & \cellcolor{lower}bad3 & \cellcolor{lower}SEE & \cellcolor{lower}EPE \\
        \hline
        \multirow{3}{*}{\rotatebox[origin=c]{90}{$\kappa=2$}} & $\mathcal{D}$\cite{hirschmuller2007stereo} & 27.56 & 24.58 & 11.99 & 23.84 \\
        & \cellcolor{ours}US & \cellcolor{ours} \textbf{21.52} & \cellcolor{ours} \textbf{16.46} & \cellcolor{ours} \textbf{5.42} & \cellcolor{ours} \textbf{5.49} \\
        & \cellcolor{ours}SF & \cellcolor{ours} 21.91 &  \cellcolor{ours} 18.01 & \cellcolor{ours} 5.63 & \cellcolor{ours} 8.13  \\
        \hline
        \multirow{3}{*}{\rotatebox[origin=c]{90}{$\kappa=4$}} & $\mathcal{D}$\cite{hirschmuller2007stereo} &  29.38 & 24.32 & 9.70 & 16.80 \\
        & \cellcolor{ours}US & \cellcolor{ours} 25.61 &\cellcolor{ours} \textbf{18.85} & \cellcolor{ours} \textbf{6.45} & \cellcolor{ours} 6.20 \\
        & \cellcolor{ours}SF &  \cellcolor{ours}\textbf{24.55} & \cellcolor{ours}19.14 & \cellcolor{ours} 6.77 & \cellcolor{ours} \textbf{6.14}  \\
        \hline

    \end{tabular}
    }
    \caption{\textbf{Generalization to Middlebury v3 - unbalanced setup.} We consider $\kappa = \{2,4\}$ and evaluate  our models trained on SceneFlow (SF) and UnrealStereo4K (US) at F. Initial disparities, $\mathcal{D}$,  are computed by SGM at H and Q  for $\kappa=2$ and $\kappa=4$, respectively.}
    \label{tab:middlebury_unbalanced}
    
\end{table}

\subsubsection{Evaluation on Middlebury v3.} Eventually, we assess the performance of our method when tested on high-res images from Middlebury v3 by simulating an unbalanced configuration. Table \ref{tab:middlebury_unbalanced} reports the results provided by neural refinement architecture  trained either on SceneFlow (SF)  or UnrealStereo4K (US), with both models run without any fine-tuning and evaluated using the available \gt{} at full resolution (F). The initial input disparities, $\mathcal{D}$,  are computed  by SGM and downsampling $\mathcal{I}_l $ by a factor $\kappa = \{2,4\}$ to match the resolution of $\mathcal{I}_r$. 
We point out that the model trained on SF is exactly the same as that used for the experiments in the balanced setting (Tabs. \ref{KITTI_2015_online_benchmark}, \ref{DRR_comparison} and \ref{synthetic-to-real}). Thus, this model is able to consistently improve the accuracy of $\mathcal{D}$ in the unbalanced setting as well, proving that the approach proposed in this paper allows for effectively addressing arbitrary resolution stereo with a single neural network trained only once. Besides, training our neural network on US, which features  higher-res images compared to SF, yields an increase in accuracy for $\kappa=2$, \ie{} when the resolution of $\mathcal{I}_r$  is closer to that of $\mathcal{I}_l$ and SGM runs on higher-resolution images. With  $\kappa=4$, conversely, training our network on either of the two datasets yields equivalent results, with both models capable of ameliorating significantly the input disparities computed by SGM. 

\section{Conclusion}
We have tackled stereo from a refinement perspective and proposed a novel, versatile neural architecture. Given as inputs an image and a disparity map, our network, trained once and solely on synthetic data, can refine it more accurately than other deep refinement approaches.  Notably, our proposal can  yield outstanding zero-shot generalization by refining disparity maps obtained by a traditional stereo matcher like SGM. In particular, we have shown superior accuracy  w.r.t. end-to-end  approaches specifically conceived to excel in out-of-domain performance.
Thanks to the proposed continuous formulation of the disparity refinement problem, our  architecture can process effectively unbalanced stereo pairs as well as predict output disparity maps at any arbitrary resolution. 

\textbf{Acknowledgements.} We gratefully acknowledge the funding support of Huawei Technologies Oy (Finland). 

{\small
\bibliographystyle{ieee_fullname}
\bibliography{egbib}
}

\newpage\phantom{Supplementary}
\multido{\i=1+1}{12}{


\section*{Supplementary material}

# Neural Disparity Refinement for Arbitrary Resolution Stereo – Supplementary Material

Filippo Aleotti*     Fabio Tosi*     Pierluigi Zama Ramirez*
Matteo Poggi     Samuele Salti     Stefano Mattoccia     Luigi Di Stefano
Department of Computer Science and Engineering (DISI)
University of Bologna, Italy
*{fabio.tosi5, filippo.aleotti2, pierluigi.zama}@unibo.it

This document provides additional details concerning our 3DV submission "Neural Disparity Refinement for Arbitrary Resolution Stereo". Firstly, we study more in detail the unbalanced stereo setting considered in the main paper by introducing a procedure to rectify stereo images captured using cameras with dramatically different properties, such as those available on a mobile phone, in Section 1, together with qualitative examples of unbalanced stereo pairs depicting real-world scenes, acquired with a custom stereo rig specifically designed to emulate smartphone camera settings and calibrated accordingly, over which we run our refinement framework. Then, in Section 2 we detail the specification of the proposed architecture and a complete description of the data augmentation used during training. In Section 3 we show a more detailed quantitative comparison with DRR [1] and finally, in Section 4, we report more qualitative results showing both disparity maps and point cloud visualizations concerning the balanced stereo setting, respectively on the SceneFlow, KITTI, Middlebury v3 and ETH3D datasets.

## 1. Calibration and Rectification of an Unbalanced Stereo Rig

In this section, we describe in detail how to accurately calibrate an unbalanced stereo rig, made of a high-resolution camera and one at lower-resolution collecting respectively frames $\mathcal{I}_l$ and $\mathcal{I}_r$. Such a rig is characterized by an unbalance factor $\kappa$, defined as the ratio between $\mathcal{I}_l$ and $\mathcal{I}_r$ widths. The calibration process allows us to *rectify* the frame pairs acquired by the rig. In such a setting, the rectification constraint shall be understood to hold up to a scale factor, *i.e.* $\mathcal{I}_l$ and $\mathcal{I}_r$ turn out to be rectified whenever resized to the same - arbitrary - shape.

We first calibrate each camera separately using the pinhole camera model. The well-known distortion-free projective transformation performed by a pinhole camera model is given by:

$$p = A\mathcal{R}\mathcal{T}P_w \quad (1)$$

where $P_w$ is a 3D point expressed w.r.t. the world reference frame (WRF), $p$ is a 2D pixel in the image plane, $A$ is the intrinsic parameters matrix and $\mathcal{R}, \mathcal{T}$ are the rotation and translation from the world reference frame (WRF) to the camera reference frame (CRF), respectively.

However, real lenses have radial and tangential distortions. We follow the lens distortion model adopted in the OpenCV library, where such a distortion is modelled through a vector of parameters $Dist = k_1, k_2, k_3, p_1, p_2$, with $k_1, k_2, k_3$ denoting the radial distortion parameters and $p_1, p_2$ the tangential distortion parameters respectively.

Given a known pattern (*e.g.*, a chessboard), we can find in the images a set of key-points (*e.g.*, inner corners of the chessboard) for which we know the exact 3D position in the WRF and, accordingly, build a set of 2D-3D correspondences which allows for inferring camera parameters through calibration. We estimate the 2D coordinates of the corners, namely $p_L, p_R$, in images acquired by $L, R$ cameras, respectively, by using a standard corner detection algorithm. By calibrating each camera of the rig independently, we estimate their intrinsic matrices $A_L, A_R$ and the lens distortion parameters $Dist_L, Dist_R$ of the $L$ and $R$ cameras, respectively. Given the intrinsic and distortion parameters, we can undistort the images to perform a stereo calibration of the stereo rig. We can thus estimate the rotations $\mathcal{R}_{LR}$ and translations $\mathcal{T}_{LR}$, from the $L$ to $R$ CRFs.

---

*Joint first authorship.

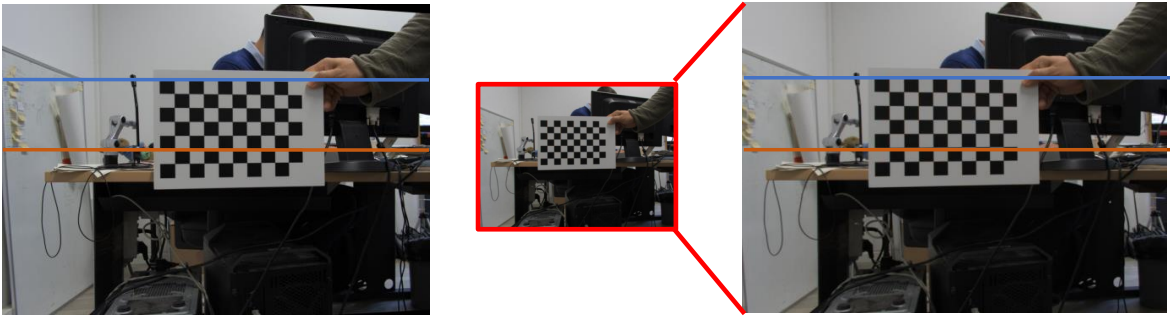

Figure 1. **Unbalanced rectified** $L$ **and** $R$ **images.** Upsampling only the low-resolution $R$ image to match the resolution of $L$ yields a rectified stereo pair at the highest resolution.

Typically when estimating a stereo rectification transformation, we assume to have both cameras at the same resolution and a similar Field of View (FOV). The new projection matrix is typically found as the mean between the initial intrinsic matrixes of the two cameras. However, as in our case, the $R$ camera has dramatically different characteristics compared to other cameras, directly performing the traditional rectification process would yield poor results. Indeed, we would need to perform harsh downsampling of the $L$ image or a large upsampling of the $R$ image to get rectified images. In contrast, we would like our rectified images to preserve their original resolution with the smallest amount of interpolation. Thus, we define the concept of *unbalanced rectification*, which allows for obtaining rectified images by performing only up-sampling or down-sampling operations, as illustrated in Fig. 1. To achieve the best possible rectification with a small amount of interpolation, we use the following procedure. First, we calculate the Horizontal Field Of Views $HFOV_L$ and $HFOV_R$ using the focal length $f_L, f_R$ (known from the intrinsic matrixes) of the $L$ and $R$ cameras, respectively. Then, we find the camera with the smaller $HFOV$, which will define an upper bound of the common visible area between the two cameras. We denote the camera with the smaller $HFOV$ as $j$ while the other one as $i$.

$$\begin{cases} i = L, j = R & \text{if } HFOV_R < HFOV_L \\ i = R, j = L & \text{if } HFOV_L < HFOV_R \end{cases} \tag{2}$$

Then, we modify the intrinsic parameters of $i$ to simulate a crop and scale of its images so as to match the $HFOV$, Aspect Ratio ($AR$) and size of $j$, and eventually calculate the rectification transformation with these parameters.

Hence, we calculate the new width and height of $i$, $\hat{W}_i$ and $\hat{H}_i$, which we use to crop the image with the larger $HFOV$ to match the smaller $HFOV$ one and to preserve the aspect ratio as follows:

$$\hat{W}_i = 2 \tan \frac{HFOV_j}{2} f_i \tag{3}$$

$$\hat{H}_i = \frac{H_j}{W_j} \hat{W}_i \tag{4}$$

Then, we modify the intrinsic parameters of $i$ to simulate the crop and resize to match the resolution of $j$ as follows:

$$\hat{A}_i = \begin{bmatrix} f_x^i \cdot \frac{W_j}{\hat{W}_i} & 0 & (u_0^i - \frac{W_i - \hat{W}_i}{2}) \cdot \frac{W_j}{\hat{W}_i} \\ 0 & f_y^i \cdot \frac{H_j}{\hat{H}_i} & (v_0^i - \frac{H_i - \hat{H}_i}{2}) \cdot \frac{H_j}{\hat{H}_i} \\ 0 & 0 & 1 \end{bmatrix}$$

We estimate the rectification transformation as we would have two cameras of height $H_j$ and width $W_j$, finding the new intrinsic $A_{L_{rect}}$ and $A_{R_{rect}}$, and the rotations $\hat{\mathcal{R}}_{Lrect}, \hat{\mathcal{R}}_{Rrect}$, of $L$ and $R$ to map the initial image plane into the rectified image plane. Finally, as we have estimated the intrinsic matrixes at the resolution of $j$, we rescale $A_{i_{rect}}$ (i.e., focal and piercing point) with a vertical and horizontal scale factors equal to $\frac{\hat{H}_i}{H_j}$ and $\frac{\hat{W}_i}{W_j}$, respectively.

Fig. 2 shows an example of raw, unbalanced stereo pair on the left, with $\mathcal{I}_l$ and $\mathcal{I}_r$ acquired respectively by two cameras at very different resolutions. The calibration procedure described in the reminder allows to rectify them, as shown on the right.

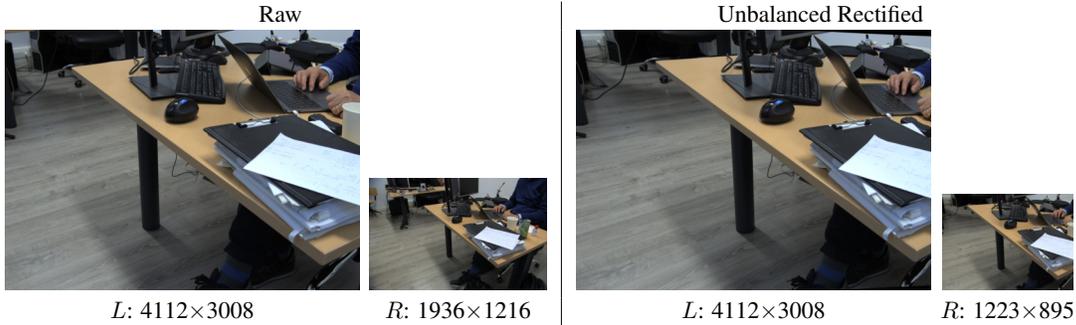
Figure 2. **Example of unbalanced stereo pair.** On left, raw images acquired by two very different cameras, respectively at $412 \times 3008$ and $1936 \times 1216$ resolution. On right, images rectified according to unbalanced calibration and rectification.

This allows us to design a custom unbalanced stereo rig, emulating the setup commonly available on mobile smartphones, that we use to collect additional samples over which we can qualitatively appreciate the effectiveness of our Neural Disparity Refinement framework. Fig. 3 shows four examples acquired in an indoor environment, framing from left to right the high-resolution image used as reference, the initial disparity map computed by means of SGM and the outcome of our network.

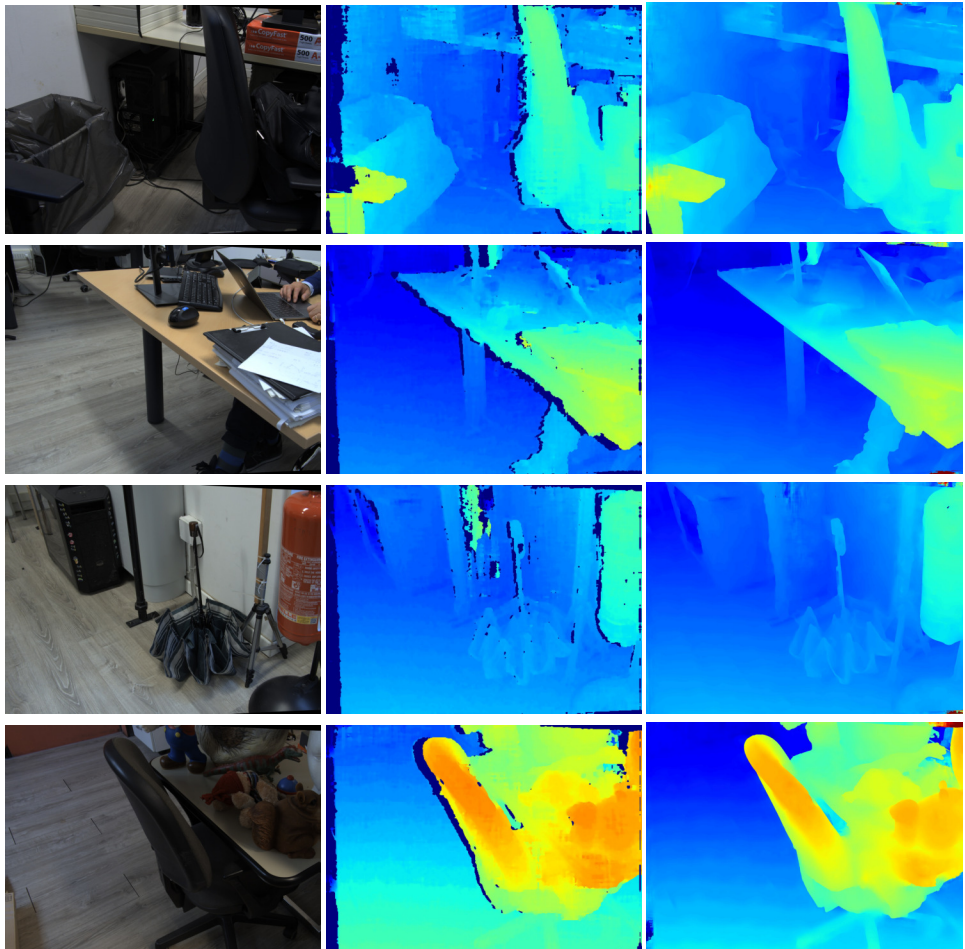
Figure 3. **Qualitative Results on a real unbalanced stereo setup**. We show qualitative results obtained by our network (trained on the synthetic SceneFlow dataset only) on real-world images captured using an unbalanced stereo setting featuring two cameras at $4112 \times 3008$ and $1223 \times 895$ resolution. From left to right, we show the high-res RGB image, the initial disparity map computed by SGM [2] and the estimated disparity at $4112 \times 3008$ resolution.

3

## 2. Implementation Details

### 2.1. Data Augmentation

We report further details of the data augmentation adopted during training as a complement of Section 4.2 of the main paper. More specifically, for both the balanced and unbalanced setups, we train our architecture by generating on the fly noisy disparity maps computed by two traditional stereo algorithms such as SGM [2] and AD-Census [6] using OpenCV. In particular, we randomly select different parameter settings for both algorithms during training. For SGM, we use $[3, 5, 7]$ as block size $b$, P1 as $2 \times b \times b$, P2 as $K \times b \times b$ where $K \in [32, 64, 96]$ and the left-right consistency check threshold as $[1, 2, 5]$ if enabled. For AD-Census, instead, we use randomly select the block size in the range $[7, 21]$ and a uniqueness ratio value in $[0, 15]$. Moreover, we adopt as input of our network a corrupted version of the ground-truth disparity by adding to it Gaussian noise and sequentially downsample and upsample it using a downsampling factor of $[1, 2, 4, 8]$. Notice that we feed our network using the input disparity computed by SGM with probability $0.45$, AD-Census with probability $0.45$ while the corrupted ground-truth with probability $0.1$. In all cases, the input disparities -and, accordingly, the ground-truth - are scaled by a factor randomly drawn from $[0.2, 3]$. We further augment the training procedure by flipping both the noisy input disparity and the RGB image horizontally and vertically. We also perform the input RGB image using several augmentations such as RGB shift, channel dropout, channel shuffle, histogram equalization, random brightness, random contrast, random gamma, motion blur, median blur, Gaussian noise, Gaussian blur, image compression and conversion to grayscale.

### 2.2. Network Architecture

In Table 1, we report a detailed description of the proposed neural refinement architecture. For each layer of the network, we show the convolution kernel size **K**, the stride **S**, the input and output number of channels and the input of the layer. Each convolutional layer of the *Encoder Disparity*, *Encoder RGB* and *Decoder* is followed by a ReLU activation function, while convolutions of both the multi-layer perceptrons are followed by the Sine activation function. The symbol "," means concatenation.

| Layer | K | S | In/Out | Input |
|---|---|---|---|---|
| **Encoder Disparity** | | | | |
| conv0_d | 3 | 1 | 1/64 | input_disparity |
| pool0_d | - | 2 | 64/64 | conv0_d |
| conv1_d | 3 | 1 | 64/64 | pool0_d |
| pool1_d | - | 2 | 64/64 | conv1_d |
| conv2_d | 3 | 1 | 64/128 | pool1_d |
| conv3_d | 3 | 1 | 128/128 | conv2_d |
| pool2_d | - | 2 | 128/128 | conv3_d |
| conv4_d | 3 | 1 | 128/256 | pool2_d |
| conv5_d | 3 | 1 | 256/256 | conv4_d |
| pool3_d | - | 2 | 256/256 | conv5_d |
| conv6_d | 3 | 1 | 256/512 | pool3_d |
| conv7_d | 3 | 1 | 512/512 | conv6_d |
| pool4_d | - | 2 | 512/512 | conv7_d |
| conv8_d | 3 | 1 | 512/512 | pool4_d |
| conv9_d | 3 | 1 | 512/512 | conv8_d |
| **Encoder RGB** | | | | |
| conv0_rgb | 3 | 1 | 3/64 | input_RGB |
| pool0_rgb | - | 2 | 64/64 | conv0_rgb |
| conv1_rgb | 3 | 1 | 64/64 | pool0_rgb |
| pool1_rgb | - | 2 | 64/64 | conv1_rgb |
| conv2_rgb | 3 | 1 | 64/128 | pool1_rgb |
| conv3_rgb | 3 | 1 | 128/128 | conv2_rgb |
| pool2_rgb | - | 2 | 128/128 | conv3_rgb |
| conv4_rgb | 3 | 1 | 128/256 | pool2_rgb |
| conv5_rgb | 3 | 1 | 256/256 | conv4_rgb |
| pool3_rgb | - | 2 | 256/256 | conv5_rgb |
| conv6_rgb | 3 | 1 | 256/512 | pool3_rgb |
| conv7_rgb | 3 | 1 | 512/512 | conv6_rgb |
| pool4_rgb | - | 2 | 512/512 | conv7_rgb |
| conv8_rgb | 3 | 1 | 512/512 | pool4_rgb |
| conv9_rgb | 3 | 1 | 512/512 | conv8_rgb |

| Layer | K | S | In/Out | Input |
|---|---|---|---|---|
| **Decoder** | | | | |
| upconv0 | 3 | 1 | 512/512 | conv9_d + conv9_rgb |
| upconv1 | 3 | 1 | 512/512 | upconv0 |
| upconv2 | 3 | 1 | 512/512 | upconv1 |
| upsample0 | - | 2 | 512/512 | upconv2 |
| upconv3 | 3 | 1 | 512/512 | upsample0 + conv7_d + conv7_rgb |
| upconv4 | 3 | 1 | 512/256 | upconv3 |
| upconv5 | 3 | 1 | 256/256 | upconv4 |
| upsample1 | - | 2 | 256/256 | upconv5 |
| upconv6 | 3 | 1 | 256/256 | upsample1 + conv5_d + conv5_rgb |
| upconv7 | 3 | 1 | 256/256 | upconv6 |
| upconv8 | 3 | 1 | 256/128 | upconv7 |
| upsample2 | - | 2 | 128/128 | upconv8 |
| upconv9 | 3 | 1 | 128/128 | upsample2 + conv3_d + conv3_rgb |
| upconv10 | 3 | 1 | 128/128 | upconv9 |
| upconv11 | 3 | 1 | 128/64 | upconv10 |
| upsample3 | - | 2 | 64/64 | upconv11 |
| upconv12 | 3 | 1 | 64/64 | upsample3 |
| upsample4 | - | 2 | 64/64 | upconv12 |
| upconv13 | 3 | 1 | 64/32 | upsample4 |
| **MLP Classification** | | | | |
| conv0_c | 1 | 1 | 96/512 | upsample3, upconv13 |
| conv1_c | 1 | 1 | 608/256 | conv0_c, upsample3, upconv13 |
| conv2_c | 1 | 1 | 352/128 | conv1_c, upsample3, upconv13 |
| conv3_c | 1 | 1 | 224/256 | conv2_c, upsample3, upconv13 |
| **MLP Offset** | | | | |
| conv0_o | 1 | 1 | 97/128 | upsample3, upconv13, argmax(conv3_c) |
| conv1_o | 1 | 1 | 225/64 | conv0_o, upsample3, upconv13 |
| conv2_o | 1 | 1 | 161/1 | conv1_o, upsample3, upconv13 |

Table 1. **Parameters of the proposed neural disparity architecture.** We report the detailed list of layers composing each of the module in our framework.

The use of two separate encoders increase the accuracy of the refined disparity maps, as we can see in Table 2. Indeed, using a single encoder processing the RGB image concatenated with the raw disparity map always lead to inferior accuracy.

| Input | Encoder | bad2 | bad3 | bad4 | bad5 | EPE | SEE |
|---|---|---|---|---|---|---|---|
| AD-Census | Single | 9.27 | 6.73 | 5.40 | 4.57 | 1.74 | 1.69 |
| | Double (Ours) | **8.49** | **6.10** | **4.86** | **4.10** | **1.53** | **1.48** |
| C-CNN | Single | 8.00 | 6.02 | 5.00 | 4.39 | 1.76 | 1.64 |
| | Double (Ours) | **7.11** | **5.25** | **4.28** | **3.68** | **1.42** | **1.36** |
| SGM | Single | 6.26 | 4.72 | 3.90 | 3.38 | 1.39 | 1.38 |
| | Double (Ours) | **5.80** | **4.33** | **3.56** | **3.07** | **1.19** | **1.26** |

Table 2. **Ablation study - single vs double encoder -** on the SceneFlow test set. We report the results obtained by our network when deploying a single or two separate encoders to process the RGB image and the raw disparity map.

| | bad2 | | bad3 | | bad4 | | bad5 | | EPE | |
|---|---|---|---|---|---|---|---|---|---|---|
| Method | Non-Occ | All | Non-Occ | All | Non-Occ | All | Non-Occ | All | Non-Occ | All |
| C-CNN [3] | 18.24 | 26.71 | 15.66 | 23.99 | 14.21 | 22.28 | 13.24 | 21.04 | 6.06 | 8.71 |
| DRR [1] | 12.85 | 17.83 | 10.10 | 14.38 | 8.29 | 11.96 | 7.06 | 10.25 | 1.77 | 2.37 |
| DRR ×2 [1] | 11.53 | 16.41 | 8.76 | 12.87 | 7.00 | 10.48 | 5.91 | 8.92 | 1.79 | 2.32 |
| Ours | **10.84** | **15.02** | **7.55** | **11.02** | **5.80** | **8.68** | **4.76** | **7.14** | **1.38** | **1.84** |

Table 3. **Stereo matching results on Middlebury v3**. Both models are trained on the SceneFlow dataset and tested on the 15 images of the Middlebury v3 training dataset at quarter resolution.

| | bad2 | | bad3 | | bad4 | | bad5 | | EPE | |
|---|---|---|---|---|---|---|---|---|---|---|
| Method | Non-Occ | All | Non-Occ | All | Non-Occ | All | Non-Occ | All | Non-Occ | All |
| C-CNN [3] | 8.83 | 10.65 | 6.41 | 8.25 | 5.22 | 7.06 | 4.51 | 6.34 | 1.70 | 2.46 |
| DRR [1] | 3.92 | 4.61 | 2.71 | 3.29 | 2.08 | 2.57 | 1.70 | 2.11 | 0.79 | 0.86 |
| DRR ×2 [1] | 3.69 | 4.28 | 2.58 | 3.08 | 2.00 | 2.42 | 1.65 | 2.00 | 0.780 | 0.84 |
| Ours | **3.30** | **3.67** | **2.27** | **2.60** | **1.79** | **2.07** | **1.5** | **1.73** | **0.75** | **0.79** |

Table 4. **Stereo matching results on KITTI 2015.** Both models are fine-tuned on 160 images of the KITTI 2015 training set and tested on the remaining 40 images of the validation set.

## 3. Additional Quantitative Results

In this section, we provide a more detailed comparison with respect to the DRR network [1]. Table 3 is provided as a complement to Table 4 of the main paper for an extensive evaluation. In particular, we evaluate both all and non-occluded regions using different threshold values for the *bad-th* metric. In our comparison, we also included the multi-iteration case for DRR (indicated as DRR×2) that indicates the results of the network after 2 iterations. Table 4, instead, shows the results obtained on the 40 validation images of the KITTI 2015 training set when both models are fine-tuned on the same 160 training images with ground-truth disparities. Similarly, it can be observed how our model significantly outperforms DRR on all metrics, thus demonstrating the superiority of our technique on the disparity refinement task.

## 4. Qualitative results

In this section, we present additional qualitative results obtained using our neural refinement network on different stereo datasets in the balanced setting. Fig. 4 shows a qualitative comparison between our model and GANet [7] on *PianoL* stereo pair from Middlebury v3, when both trained on the SceneFlow dataset. We can appreciate our our refinement network produces fewer errors and achieves an lower bad2 overall score, thus better generalizing to real images.

| RGB | GANet (bad2: 9.86%) | | Ours (bad2: 5.27%) | |
|---|---|---|---|---|

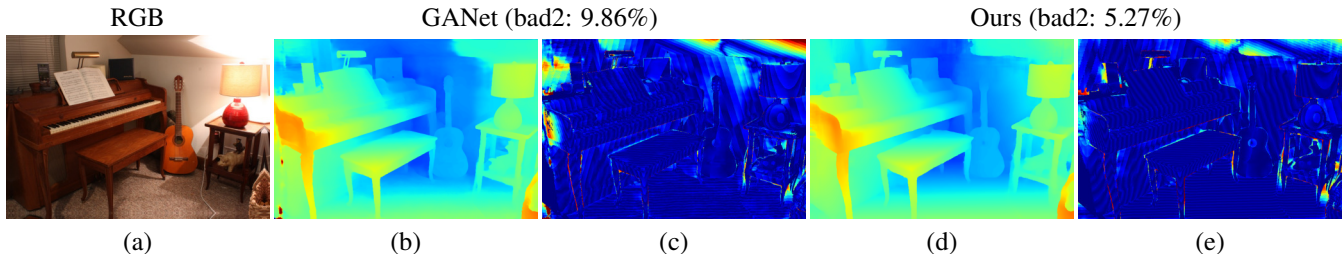

| (a) | (b) | (c) | (d) | (e) |

Figure 4. **Qualitative comparison with GANet [7].** From left to right: reference image (a), disparity and error maps by GANet (b,c) and our method (d,e).

Fig. 5 collects some examples from the SceneFlow testing split, showing from left to right the reference image, the raw disparity map estimated by SGM and the final output by our network. Figures 6, 7 and 8 report additional qualitative results

5

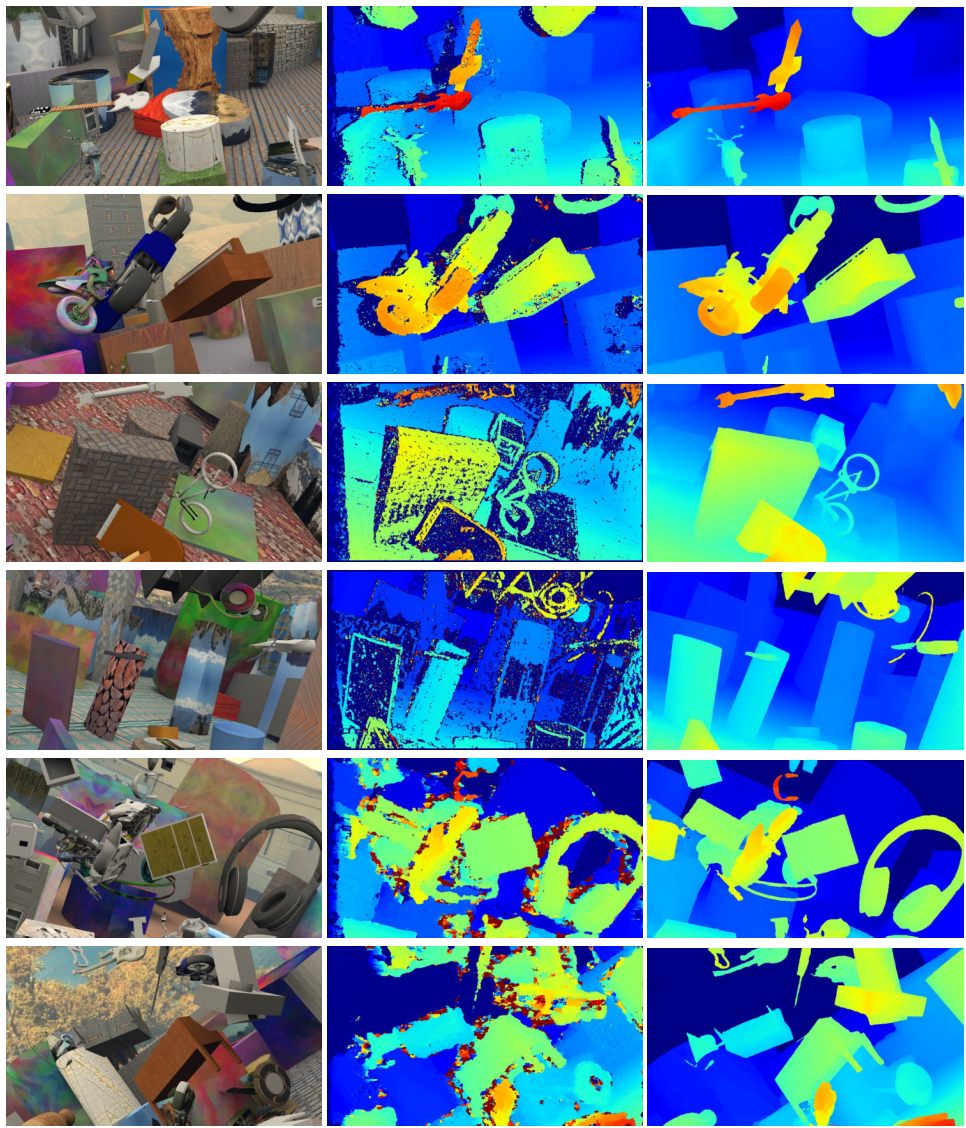

Figure 5. **Qualitative results on the SceneFlow Test Set.** We report qualitative results of our neural disparity refinement network on the SceneFlow testing set. From left to right, the RGB input image, the noisy input disparity map computed by SGM [2] (rows 1-2), AD-Census [6] (rows 3-4), C-CNN [3] (rows 5-6) and the corresponding refined disparity estimated by our network.

on real datasets, respectively KITTI 2015, Middlebury 2014 and ETH3D, highlighting once again the outstanding zero-shot generalization performance achieved by our network, trained on synthetic datasets only.

To better appreciate the sharpness of the disparity maps predicted by our formulation, in Fig. 9 we show 3D pointclouds obtained from disparity maps estimated, from top to bottom, respectively by HSMNet [5], AANet [4] and GANet [7]. We can notice how refining their predictions we remove all the flying pixels introduced by the original, over-smoothed predictions. Finally, in Fig. 10 we show a comparison between disparity maps yielded by our continuous formulation with respect to what was obtained through nearest-neighbour interpolation, highlighting once more the finer details produced by our model when dramatically increasing the resolution up to 80Mpx.

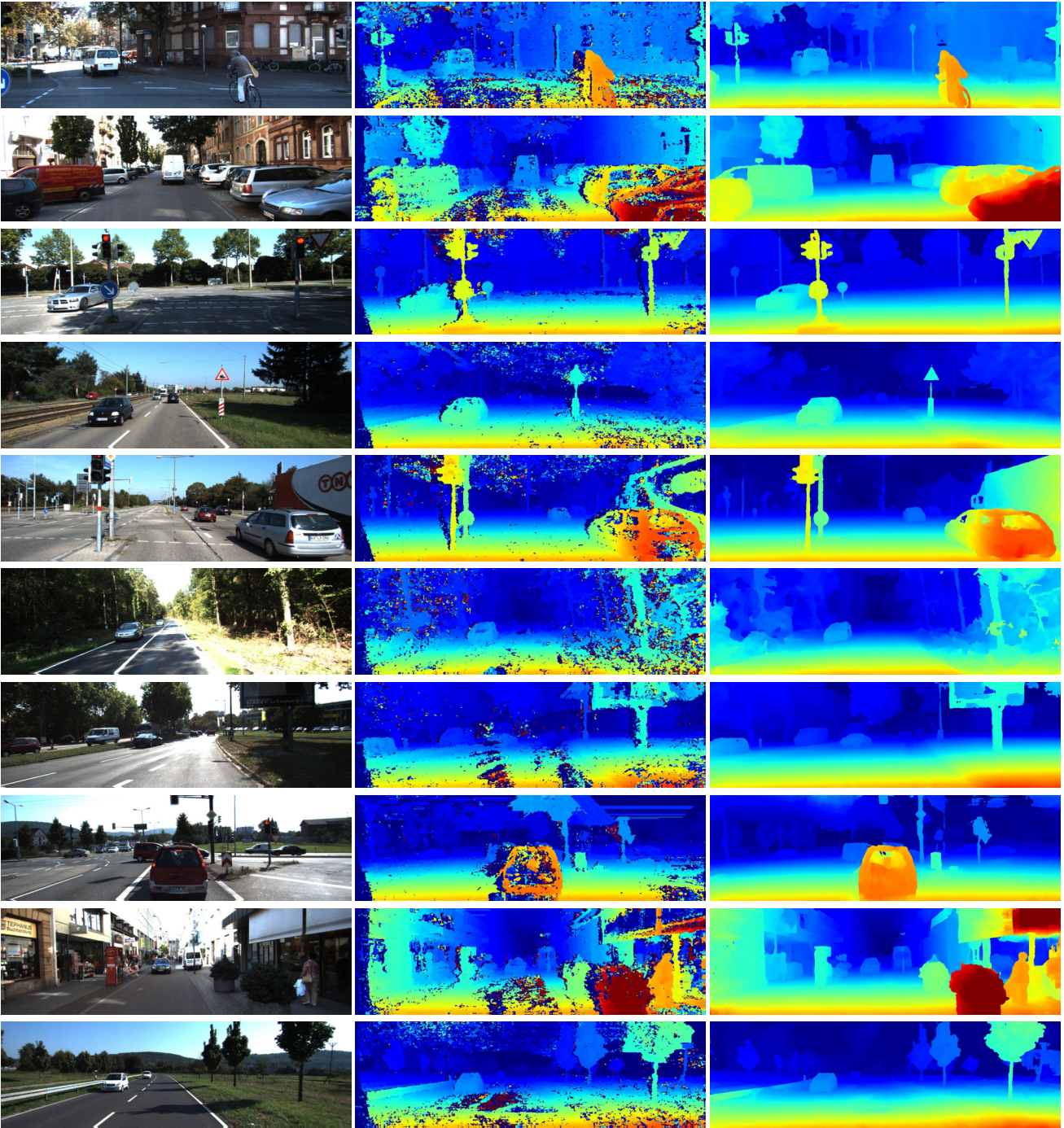

Figure 6. **Qualitative results on the KITTI 2015 Training Set**. Here, we show qualitative results concerning the generalization capability of our network (pre-trained on SceneFlow) on the KITTI 2015 training set. From left to right, the RGB input image, the noisy input disparity map computed by SGM [2] and the refined disparity estimated by our network.

7

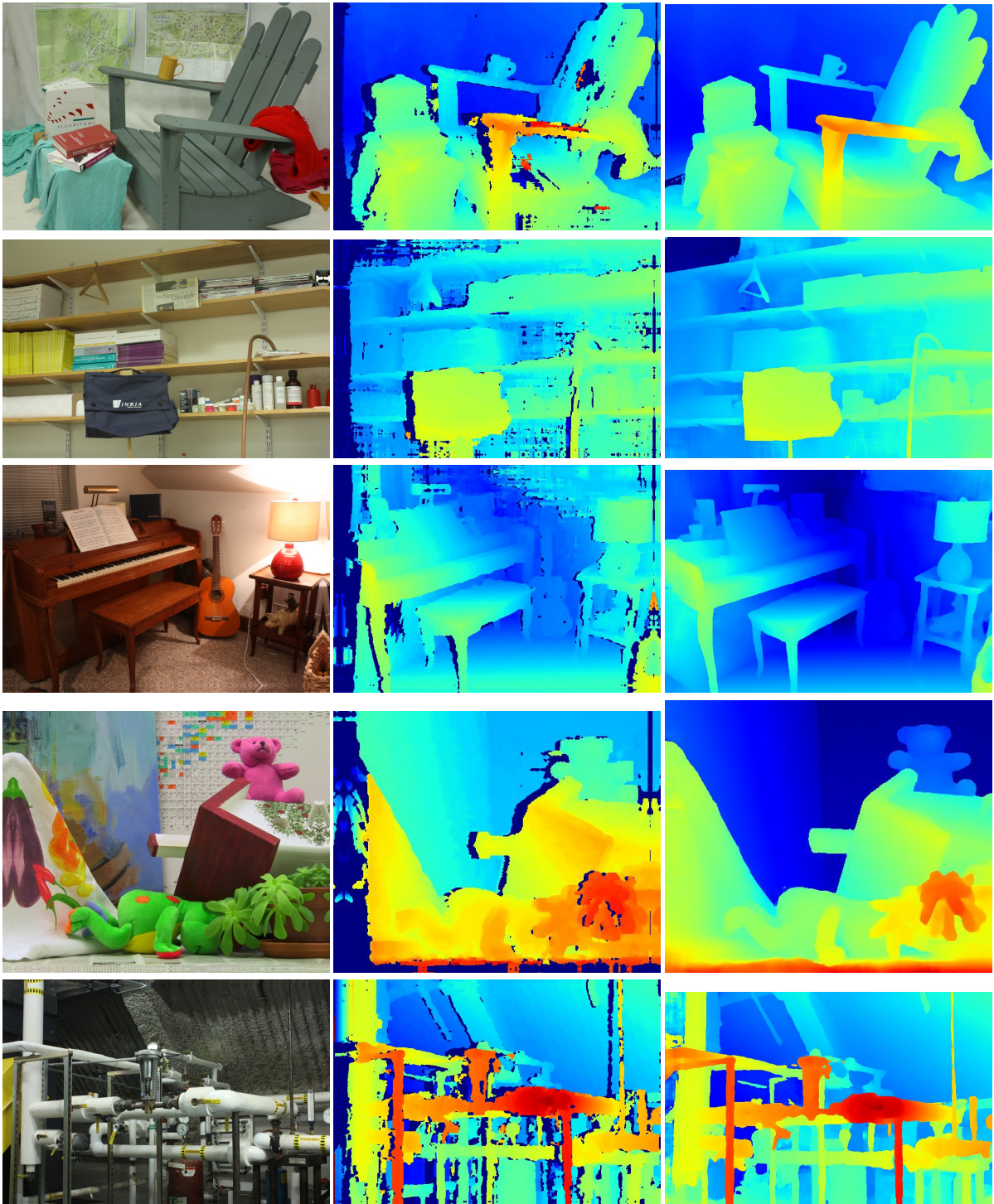

Figure 7. **Qualitative results on the Middlebury v3 Training Set**. Here, we show qualitative results concerning the generalization capability of our network (pre-trained on SceneFlow) on the Middlebury v3 training set. From left to right, the RGB input image, the noisy input disparity map computed by SGM [2] and the refined disparity estimated by our network.

8

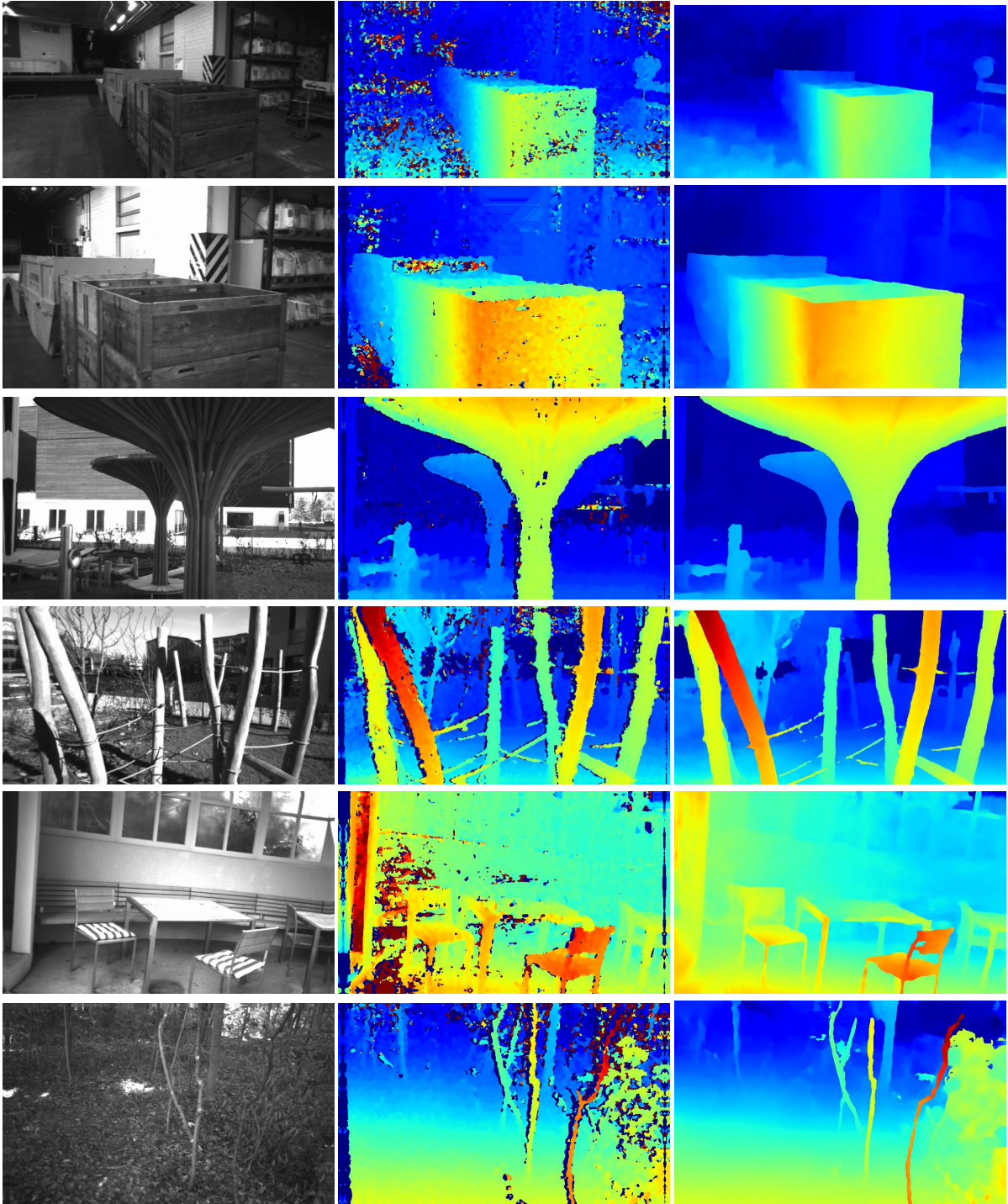

Figure 8. **Qualitative results on the ETH3D Training Set**. Here, we show qualitative results concerning the generalization capability of our network (pre-trained on SceneFlow) on the ETH3D training set. From left to right, the RGB input image, the noisy input disparity map computed by SGM [2] and the refined disparity estimated by our network.

9

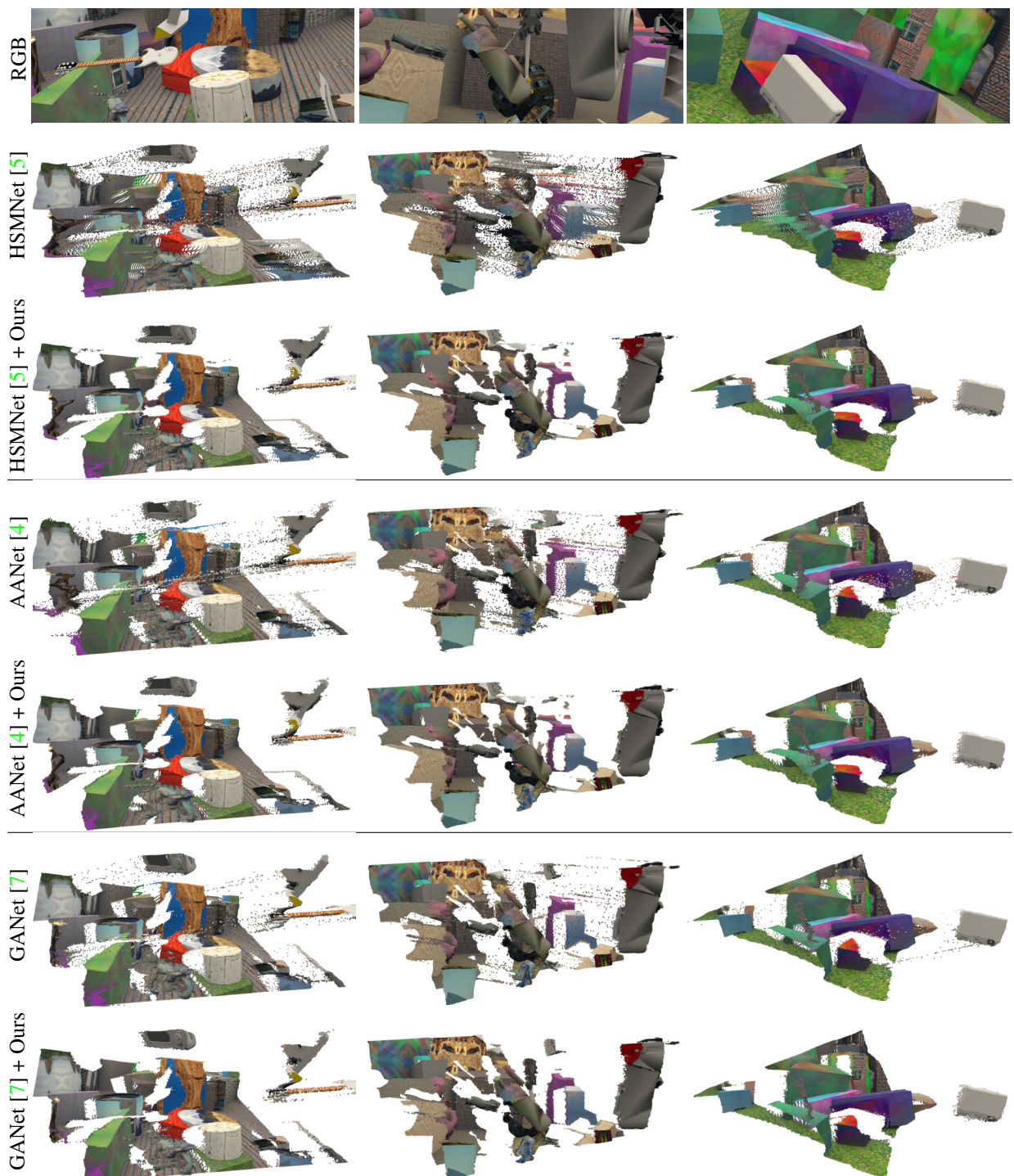

Figure 9. **Point Cloud Comparison** on the SceneFlow dataset. We show the outcomes of different state-of-the-art deep stereo networks and the point clouds obtained using our refinement method on the initial disparity estimates. Note how our network allows us to notably alleviate the bleeding effect at edge boundaries, thus resulting in more accurate 3D reconstructions. Please zoom in for details.

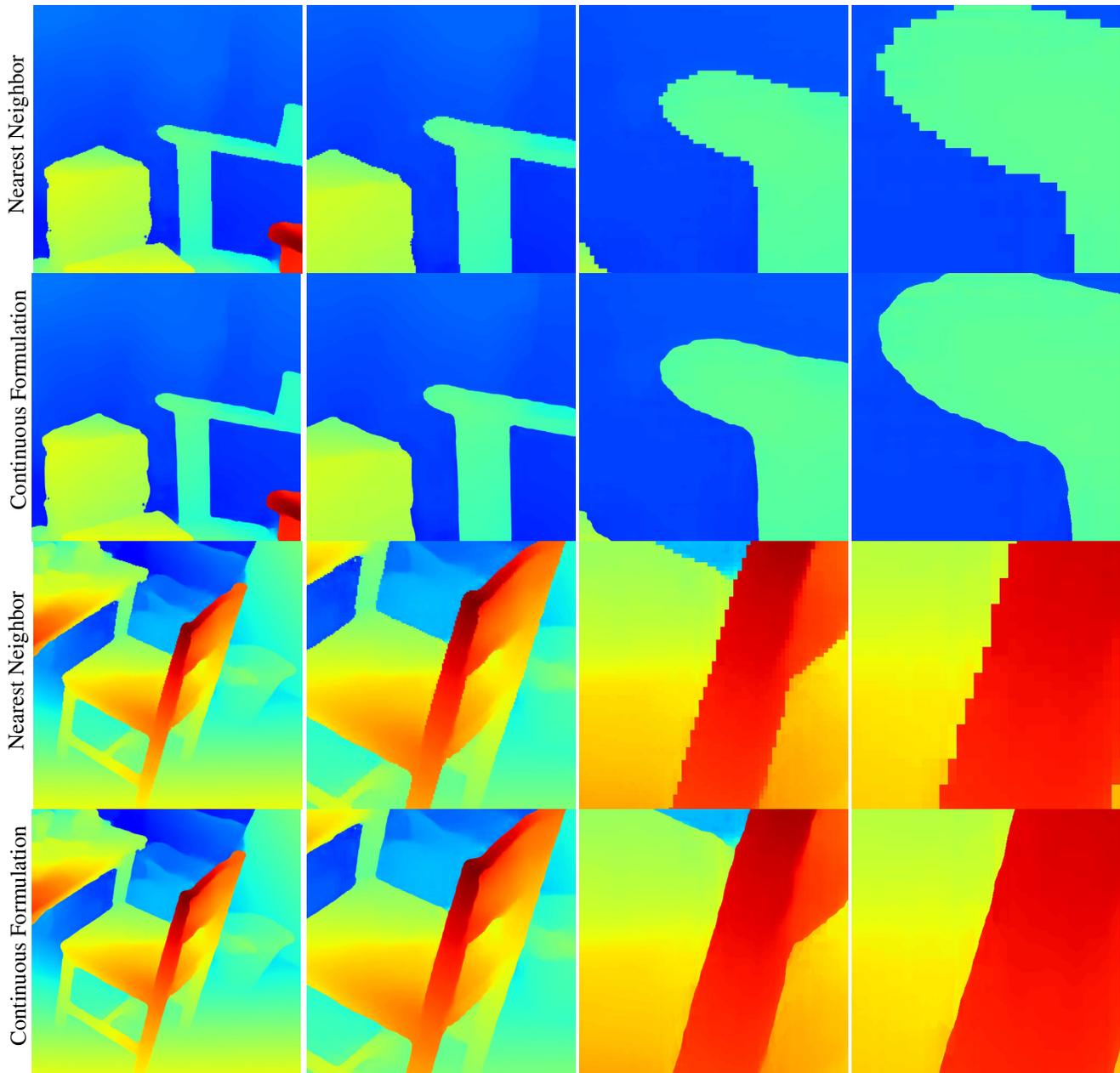

Figure 10. **Upsampling comparison**. Here, we qualitatively show the effectiveness of our continuous formulation compared to the traditional nearest-neighbor interpolation. In particular, given two noisy disparity maps at low resolution (∼ 0.3 Mpx) as input of our network, we refine and upsample them by adopting an upsampling factor of 16 (∼ 80 Mpx). It can be observed how our formulation allows us to obtain more precise disparities at object boundaries.


}

\end{document}